%% file: main.tex
\documentclass[runningheads]{llncs}

\usepackage[mobile]{eccv}

\usepackage{eccvabbrv}

\usepackage{graphicx}
\usepackage{booktabs}
\usepackage{ulem} 
\usepackage{capt-of}
\usepackage[accsupp]{axessibility}  % Improves PDF readability for those with disabilities.
\usepackage{caption}

\usepackage[pagebackref,breaklinks,colorlinks,citecolor=eccvblue]{hyperref}
\usepackage{orcidlink}

\usepackage{booktabs}
\usepackage{xcolor}
\usepackage{multirow}
\usepackage{tabularx}
\newcolumntype{C}{>{\centering\arraybackslash}X}
\definecolor{light-yellow}{rgb}{1.0, 0.77, 0.06}
\usepackage{color} 
\usepackage{colortbl}
\usepackage{graphicx}
\usepackage{xcolor}
\usepackage{makecell}
\usepackage[most]{tcolorbox} % 加载并启用所有高级特性
\definecolor{blackish}{rgb}{0.2, 0.2, 0.2} % 自定义你框框用的颜色
\definecolor{beaublue}{rgb}{0.74, 0.83, 0.9}
\definecolor{lightgraybg}{rgb}{0.95, 0.95, 0.95} % 新增的淡灰色
\definecolor{red}{rgb}{1.0,0.0,0.0}
\definecolor{green}{rgb}{0.0,1.0,0.0}
\definecolor{blue}{rgb}{0.0,0.0,1.0}
\definecolor{darkpastelpurple}{rgb}{0.59, 0.44, 0.84}

\begin{document}

\title{LiveWorld: Simulating Out-of-Sight Dynamics in Generative Video World Models}
%\title{LiveWorld: Formulating Event Permanency in Generative World Modeling} 
% \title{Turn Away, Time Continues: Out-of-Sight Object Permanence for Video World Models}

\titlerunning{LiveWorld}

\author{
Zicheng Duan$^{1*}$,
Jiatong Xia$^{1*}$,
Zeyu Zhang$^{2*}$,
Wenbo Zhang$^{1}$,
Gengze Zhou$^{1}$,
Chenhui Gou$^{3}$,
Yefei He$^{4}$,
Feng Chen$^{1\dagger}$,
Xinyu Zhang$^{5\ddagger}$,
Lingqiao Liu$^{1\dagger\ddagger}$
}

\authorrunning{Z. Duan et al.}

\institute{
$^{1}$Adelaide University, $^{2}$The Australian National University, $^{3}$Monash University,
$^{4}$Zhejiang University, $^{5}$University of Auckland
\\
\email{zicheng.duan@adelaide.edu.au, lingqiao.liu@adelaide.edu.au}
\\
{\small * Equal contribution. \quad $\dagger$ Project lead. \quad $\ddagger$ Corresponding author.}
}

\maketitle

\input{secs/0_abstract}
\input{secs/1_introduction}
\input{secs/2_related_work}
\input{secs/3_method}
\input{secs/4_experiments}

\bibliographystyle{splncs04}
\bibliography{main}

\newpage
\appendix
\input{secs/5_appendix}

\end{document}

%% file: secs/0_abstract.tex
% \begin{figure}[t]
%     \centering
%     \includegraphics[width=1\linewidth]{imgs/teaser.jpg}
%     \caption{\textbf{LiveWorld enables persistent out-of-sight dynamics.} Instead of freezing unobserved regions, our framework explicitly decouples world evolution from observation rendering. We register stationary \textit{Monitors} to autonomously fast-forward the temporal progression of active entities (e.g., the dog and the person) in the background. As the observer explores the scene along the target trajectory (green cameras), our state-aware renderer projects the continuously evolved world states to synthesize the final observation. This ensures that dynamic events progress naturally, accurately reflecting the elapsed time even when entities are completely out of the observer's view.}
%     \label{fig:placeholder}
% \end{figure}

% \maketitle

\begin{center}
    \vspace{-5mm}
    \centering
    \includegraphics[width=\linewidth]{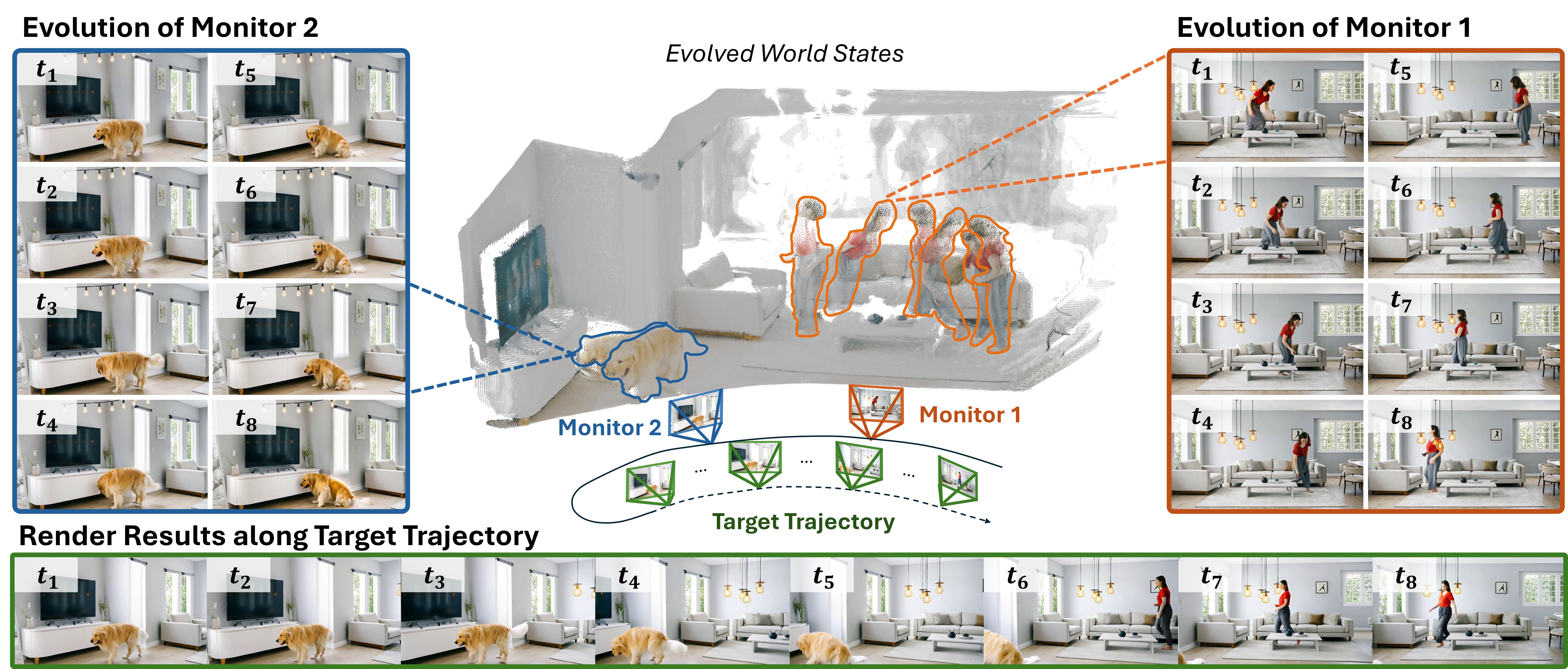}
    \vspace{-5mm}
    \captionof{figure}{\textbf{LiveWorld enables persistent out-of-sight dynamics.} Instead of freezing unobserved regions, our framework explicitly decouples world evolution from observation rendering. We register stationary \textit{Monitors} to autonomously fast-forward the temporal progression of active entities (e.g., the dog and the person) in the background. As the observer explores the scene along the target trajectory (green cameras), our state-aware renderer projects the continuously evolved world states to synthesize the final observation. This ensures that dynamic events progress naturally, accurately reflecting the elapsed time even when entities are completely out of the observer's view.}
    \label{fig:teaser}
    % \vspace{-2mm}
\end{center}

\begin{abstract}
\vspace{-2mm}

Recent generative video world models aim to simulate visual environment evolution, allowing an observer to interactively explore the scene via camera control. However, they implicitly assume that the world only evolves within the observer's field of view. Once an object leaves the observer’s view, its state is ``frozen'' in memory, and revisiting the same region later often fails to reflect events that should have occurred in the meantime. In this work, we identify and formalize this overlooked limitation as the ``out-of-sight dynamics'' problem, which impedes video world models from representing a continuously evolving world. To address this issue, we propose LiveWorld, a novel framework that extends video world models to support persistent world evolution. Instead of treating the world as static observational memory, LiveWorld models a persistent global state composed of a static 3D background and dynamic entities that continue evolving even when unobserved. To maintain these unseen dynamics, LiveWorld introduces a monitor-based mechanism that autonomously simulates the temporal progression of active entities and synchronizes their evolved states upon revisiting, ensuring spatially coherent rendering. For evaluation, we further introduce LiveBench, a dedicated benchmark for the task of maintaining out-of-sight dynamics. Extensive experiments show that LiveWorld enables persistent event evolution and long-term scene consistency, bridging the gap between existing 2D observation-based memory and true 4D dynamic world simulation. 
The baseline and benchmark will be publicly available.

\vspace{-0.5mm}
\end{abstract}

%% file: secs/1_introduction.tex
% \vspace{-8mm}
\section{Introduction}
% \vspace{-1mm}
Recently, there is an increasing demand to build world models that can anticipate future world states based on the current context and control inputs. By modeling the underlying dynamics of the environment, such systems can simulate the progression of a virtual world, offering a powerful platform for applications including agent training\cite{pan,nwm,peva}, decision-making\cite{waymo,magicdrive}, and large-scale synthetic environment generation\cite{genie1,genie3}.

% 1. 背景：Video World Model 很火，但存在问题
Among them, generative video models have emerged as a dominant paradigm for world modeling, leveraging their powerful prior to simulate realistic visual dynamics and enabling users to explore virtual environments through camera control. To maintain temporal consistency during exploration, existing approaches typically condition the generation on historical contexts stored as 2D snapshots in a KV cache~\cite{selfforcing, longlive} or explicitly reconstructed 3D spatial memory~\cite{spatia, worldplay}.

% 2. 核心痛点：Out-of-sight dynamics (结合狗吃狗粮的生动例子)
Despite these advancements, current video world models suffer from a fundamental limitation: they conflate the autonomous evolution of the world with camera-dependent rendering. By implicitly collapsing these two distinct processes into a single black-box generator, they are inherently trapped in a \textit{static-world assumption}. Consequently, once an active entity leaves the observer's field of view, its temporal progression is entirely ignored, effectively freezing the entity at its last observed timestamp. We refer to this overlooked phenomenon as the missing of \textbf{out-of-sight dynamics}. For instance, if an observer looks away from a dog eating its food, revisiting the same location later will simply retrieve the outdated snapshot of the dog mid-bite, rather than reflecting the elapsed dynamics of it having finished the meal.

% 3. 我们的洞察和解法：解耦 + 结构化近似
To overcome this limitation, we propose \textbf{LiveWorld}, a novel video world model framework that explicitly decouples world evolution ($\mathcal{E}$) from observation rendering ($\mathcal{R}$). Recognizing that maintaining a fully dense 4D state of the entire unobserved world is computationally intractable, we introduce a \textit{structured world-state approximation}. We factorize the global world state into two components based on their physical nature: a temporally invariant static background ($\mathcal{M}_{static}$), which is accumulated into a 3D spatial point cloud; and sparsely distributed dynamic entities ($\mathcal{M}_{dyn}$), which explicitly retain their temporal dimensions to continue evolving out of sight. 

% 4. 系统落地：Monitor + Unified Backbone
To seamlessly maintain and evolve this decoupled 4D world, we design a \textit{monitor-driven pipeline}. When a dynamic entity is detected, the system retroactively registers a virtual ``Monitor'' at its location. Even when the entity is no longer in the observer's field of view, the monitor autonomously fast-forwards its temporal progression, yielding an evolving 4D dynamic point cloud. To render the observer's continuous view, we project both the static environment and the up-to-date dynamic entities onto the target camera trajectory to serve as precise geometric conditioning. Importantly, recognizing that both unobserved temporal evolution and world rendering share the same generative paradigm, we implement both the monitor and the renderer using a \textit{unified state-conditioned video diffusion backbone} that takes the projected states and auxiliary references as input to produce the rendering results and the evolution video. In summary, our main contributions are as follows:

% \vspace{-0.8mm}
\begin{itemize}
    \item We rigorously identify and formalize the missing \textit{out-of-sight dynamics} problem in current video world models, highlighting the critical flaw of conflating world evolution with observation rendering.
    \item We propose \textbf{LiveWorld}, a decoupled video world model framework featuring a monitor-centric evolution system and a unified video backbone, which enables the autonomous temporal progression of unobserved entities.
    \item We develop the first dedicated benchmark, \textbf{LiveBench}, specifically designed to quantitatively evaluate long-horizon out-of-sight dynamics and event permanence for video world models.
    \item Extensive experiments on LiveBench demonstrate that LiveWorld successfully bridges the gap between 2D static memorization and 4D dynamic simulation, significantly outperforming existing baselines.
\end{itemize}

%% file: secs/2_related_work.tex
\vspace{-2mm}
\section{Related Works}
\vspace{-1mm}
\subsection{Video World Models}
% \vspace{-2mm}
\label{sec:related_world_modeling}
World modeling aims to construct an evolving environment to simulate the real world. Existing works approach this problem from different perspectives. \cite{marble} builds explicit geometric-consistent 3D representations, while JEPA \cite{vjepa, vjepa2,geoworld} learns abstract state transitions in latent space. With the rapid progress of video generation models~\cite{svd,animatediff,wan,hunyuanvideo}, Generative Video World Models \cite{genie1,genie3,waymo} have become a dominant paradigm as they achieve more scalable and realistic world modeling. Typically, a video world model predicts future frames conditioned on historical context and control signals. First, to leverage historical content, early methods \cite{svd,animatediff,wan,hunyuanvideo,history} concatenate anchor frames at the sequence start. CausVID \cite{CausVID} pioneeringly distills fixed-size self-attention into causal attention, enabling autoregressive next-frame prediction with KV cache \cite{selfforcing, selfforcing++,longlive,rollingforcing,deepforcing,blockvid,inferix,contextasmemory,genie1,genie3,liveavatar} to store arbitrary history tokens. Next, to support controllable exploration, recent works \cite{hunyuangamecraft,yan,pan,relic,matrixgame, cameractrl1,cameractrl2} incorporate camera trajectory embeddings. Combined with cached history, these methods allow revisiting previously observed regions. However, the cached frame tokens merely capture 2D visual snapshots of past regions at the exact time they were observed. This inherently relies on a static-world assumption, completely failing to maintain out-of-sight dynamics.

\vspace{-1.5mm}
\subsection{Explicit Spatial Memory}
To achieve precise camera control and maintain long-term geometry during exploration, recent methods \cite{spatia, vmem,hunyuanworld,hunyuanworld15,worldplay,spatialmem,epic,anchorweave,code2worlds} have been developed conditioned on explicit 3D spatial memory. These approaches maintain an explicitly reconstructed spatial representation as a form of global memory---such as a point cloud combined with camera parameters predicted by feed-forward estimators \cite{wang2025vggt,wang2024dust3r,wang2025continuous,depthanything3,keetha2026mapanything,stream3r2025,tid3r}. By injecting this structural representation into the video generation model, they ensure rigorous geometric consistency.
Despite these advancements in spatial tracking, the temporal dimension of historical locations remains neglected. The registered 3D representations store merely the static 3D spatial structure of the scene at the moment of observation. Consequently, they still cannot capture the \textit{out-of-sight dynamics} within previously visited areas.

\vspace{-2.5mm}
\subsection{Out-of-sight Dynamics}
\vspace{-1mm}
An ideal world model should capture the continuous evolution of the entire explored space, forming a unified 4D spatio-temporal field where scene states evolve consistently over time. This poses a fundamental challenge to existing video world models. Current approaches~\cite{selfforcing, selfforcing++,longlive,rollingforcing,deepforcing,blockvid, contextasmemory,genie1,genie3,liveavatar, spatia, vmem,hunyuanworld,hunyuanworld15,worldplay,spatialmem,epic,anchorweave} only update states within the camera’s visible region, while content outside the current view is merely stored as historical observations in memory (e.g., KV cache or spatial memory), remaining frozen at their last observed timestamps. In this work, we formalize this overlooked problem within the video world model paradigm. Building upon explicit spatial memory, we propose a decoupled framework to continuously maintain and evolve this out-of-view representation, thereby bridging the gap between static scene memorization and true 4D dynamic world simulation.

%% file: secs/3_method.tex
\begin{figure}[t!]
    \vspace{-1mm}
    \centering
    \includegraphics[width=1\linewidth]{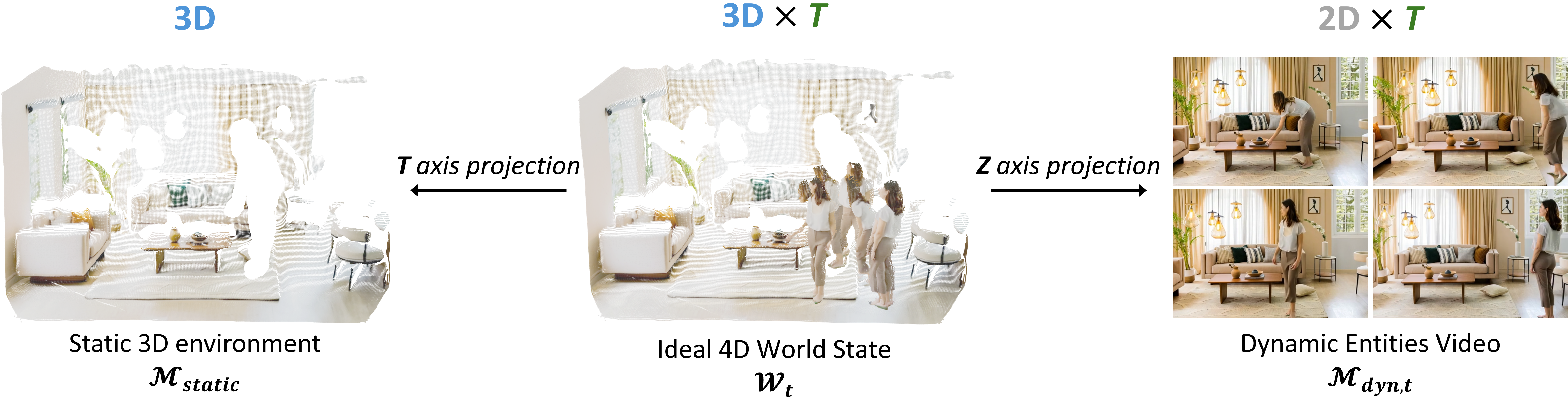}
    \vspace{-6.5mm}
    \caption{\textbf{World State Formulation.} We approximate the intractable 4D world state $\mathcal{W}_t$ by decoupling it into two trackable representations: a temporally-invariant static 3D environment $\mathcal{M}_{static}$ via $T$-axis projection, and 2D video sequences of dynamic entities $\mathcal{M}_{dyn,t}$ via $Z$-axis projection.}
    \label{fig:problem_formulation}
    \vspace{-4.5mm}
\end{figure}

\vspace{-2.5mm}
\section{Methods}
\vspace{-1.5mm}

We propose \textcolor{blue}{\textbf{LiveWorld}}, a video world model framework designed to maintain and continuously evolve the out-of-sight dynamics of the currently unobserved world. Unlike previous video world models that follow an observer-centric paradigm---conflating world evolution with rendering and freezing out-of-view regions into static snapshots---our system explicitly decouples temporal evolution from spatial rendering. By implementing a monitor-centric pipeline, we autonomously model the temporal progression of active entities, narrowing the gap between the 2D video world modeling and the 4D dynamic world.

\vspace{-2.5mm}
\subsection{Problem Formulation}
\vspace{-1.5mm}
\subsubsection{Decoupling World Evolution from Observation Rendering.}
\label{sec:problem_formulation}
The world continues to evolve even when it is not observed. That is to say, an ideal world model should maintain a latent global world state $\mathcal{W}_{t}$ at each time step $t$, which specifies the underlying 4D scene at that moment. Since this state is view-independent, while an observed frame is only a camera-dependent projection, world modeling naturally decomposes into two processes: \textbf{(1) state evolution}, which updates the world over time, and \textbf{(2) rendering}, which maps the current state to an observation under a condition $C_t = \{C_t^{\text{cam}}, C_t^{\text{text}}\}$ (where $C_t^{\text{cam}}$ is the camera pose and $C_t^{\text{text}}$ is the semantic prompt). Formally,
\vspace{-1mm}
\begin{equation}
    \mathcal{W}_{t} = \mathcal{E}(\mathcal{W}_{<t}), \quad
    F_t = \mathcal{R}(\mathcal{W}_{t}, C_t).
    \label{eq:ideal_formulation}
\end{equation}
where $\mathcal{E}$ denotes the evolution engine and $\mathcal{R}$ denotes the conditioned renderer.

However, existing video world models do not maintain such an explicit world state. Instead, they compress the evolving 4D world into a history of 2D observations and directly predict the next frame from previously observed frames $\text{F}_{<t}=\left \{ F_{t-1}, F_{t-2},\dots ,F_0 \right \} $ and control signals $C_t$:
\begin{equation}
    F_t = \mathcal{V}_{\theta}(\text{F}_{<t}, C_t).
    \label{eq3:current_formulation}
\end{equation}

Under this formulation, world evolution and rendering are implicitly collapsed into a single black-box generator $\mathcal{V}_{\theta}$. The key limitation is that $\text{F}_{<t}$ contains only camera-dependent visual snapshots rather than the full underlying state $\mathcal{W}_{<t}$. In other words, the continuous 4D world is flattened into a sequence of 2D observations, and once a region leaves the field of view, the model has no explicit state to update, so its temporal progression is ignored and the region remain frozen at its last observed timestamp. We refer to this missing temporal progression as \textbf{out-of-sight dynamics}. 
% For example, if a dog is observed eating in $F_{t-k}$, the model stores only that observed appearance; when the camera revisits the same location at time $t$, it retrieves this outdated snapshot instead of reflecting the elapsed dynamics.

\subsubsection{Structured World-State Approximation}
To address this, \textbf{LiveWorld} restores an explicit separation between \textit{world evolution} and \textit{camera-conditioned rendering}. Directly maintaining a fully explicit 4D world state is impractical; therefore, we structurally approximate $\mathcal{W}_{t}$ based on a simple intuition: the static scene is temporally invariant, while temporal changes are concentrated in sparse dynamic entities.
Accordingly, we decompose the world state into two components as illustrated in Fig.~\ref{fig:problem_formulation}. \textbf{(1) Static background.} We collapse the time-invariant background of the world along the temporal axis into a static 3D scene representation $\mathcal{M}_{static}$. For \textbf{(2) dynamic entities}, their states continuously evolve over time. To maintain their up-to-date representations $\mathcal{M}_{dyn,t}$ (especially when they are out of the observer's sight), we introduce an explicit evolution function $G_{\theta}^{\text{evo}}$ serving as $\mathcal{E}$ in Eq.~\ref{eq:ideal_formulation}. It takes historical frames $\text{F}_{<t}$ containing active entities and simulates their continuous temporal progression, yielding the evolved dynamic representation at time $t$:
\vspace{-1mm}
\begin{equation}
    \mathcal{M}_{dyn,t} = G_{\theta}^{\text{evo}}(\text{F}_{<t})
    \label{eq4:dynamic_memory}
\end{equation}
\vspace{-1mm}
The up-to-date world state is then approximated as:
\begin{equation}
    \mathcal{W}_{t} \approx \{\mathcal{M}_{static}, \mathcal{M}_{dyn,t}\}
    \label{eq:world_state}
\end{equation}
Under this formulation, the video generator no longer implicitly handles hidden temporal evolution. Instead, it serves strictly as a state-aware renderer $G_{\theta}^{\text{render}}$ (serving as $\mathcal{R}$ in Eq.~\ref{eq:ideal_formulation}) that projects the composed world state under the control signal $C_t$:
\begin{equation}
    F_t = G_{\theta}^{\text{render}}(\mathcal{W}_{t}, C_t).
    \label{eq5:our_formulation}
\end{equation}

Implementing this decoupled formulation requires two components: (1) an explicit evolution mechanism $G_{\theta}^{\text{evo}}$ to update unobserved active regions over time (Sec.~\ref{sec:event_evolution}), and (2) a state-aware generative process $G_{\theta}^{\text{render}}$ that renders the composed 3D/2D world state into a coherent observation (Sec.~\ref{sec:observer_rendering}).

\vspace{-1mm}
\subsection{Unified State-Conditioned Video Backbone}
\label{sec:shared_video_backbone}
While the state-aware renderer $G_{\theta}^\text{render}$ and the evolution engine $G_{\theta}^\text{evo}$ are conceptually distinct, they share a fundamental generative paradigm: both synthesize future visual content conditioned on previous world states and external control signals. Motivated by this structural commonality, we propose a unified state-conditioned video diffusion model $G_{\theta}$ as an abstract interface model. Moreover, although we formulate state evolution at an atomic time step $t$ in Sec.~\ref{sec:problem_formulation}; In practice, we follow foundational video diffusion models\cite{wan, svd, hunyuanvideo} that generate video chunks in discrete autoregressive rounds. Therefore, we instantiate the generative step to cover a temporal window of $T$ frames. In the following sections, we use the notation $t:t+T$ to denote the generation of a $T$-frame sequence starting from timestamp $t$. Specifically, $G_{\theta}$ comprises a latent Video Diffusion Transformer (DiT)~\cite{wan} backbone, augmented with a dual-injection conditioning design to process explicit state projections and detailed appearances respectively:
% \noindent
% \vspace{-3mm}
\paragraph{Explicit state conditioning via state adapter.} To inject the maintained world states into the generation process, we employ a \textit{state adapter} initialized from \cite{vace}. This module functions as a ControlNet~\cite{controlnet}, taking a pixel-level projection tensor $\mathbf{P}_{t:t+T} \in \mathbb{R}^{T \times H \times W \times C}$ that represents the explicitly projected state for the target generation window. By injecting these signals into the DiT backbone, the state adapter imposes strict, explicit pixel-level guidance for the generated frames, ensuring consistency with the underlying world state.

% \noindent
% \vspace{-5mm}
\paragraph{Incorporating complementary appearance references.} Because the projected state $\mathbf{P}_{t:t+T}$ primarily serves as structural and positional guidance and may lack fine-grained visual details, inspired by \cite{iclora,spatia,catvton}, we register learnable LoRA parameters to the DiT backbone to accept concatenated historical reference frames. These references typically include a temporal anchor (e.g., the immediately preceding frames) to maintain motion continuity, and appearance anchors (retrieved from past frames) to fill in dense visual textures. We concatenate these reference frames along the token axis to the input of the DiT backbone. By flexibly configuring the input projection $\mathbf{P}_{t:t+T}$ and the appearance references, this unified backbone can be seamlessly instantiated for entirely different roles. As detailed below, we utilize this shared architecture to perform autonomous unobserved evolution (Sec.~\ref{sec:event_evolution}) and observer-centric spatial rendering (Sec.~\ref{sec:observer_rendering}).

\begin{tcolorbox}[width=1.0\linewidth, colframe=blackish, colback=lightgraybg, boxsep=0mm, arc=1mm, left=1mm, right=1mm, top=1mm, bottom=1mm]
\textbf{Unified Backbone Interface.} The shared architecture $G_{\theta}$ synthesizes a $T$-frame video chunk $V_{t:t+T}$, guided by three abstract conditioning inputs: an explicit state projection $\mathbf{P}$ (which encapsulates the geometric camera control $C^{\text{cam}}$), supplementary appearance references $\mathbf{A}$, and the text prompt $C^{\text{text}}$.
\begin{equation}
    V_{t:t+T} = G_{\theta}(\mathbf{P}_{t:t+T},\, \mathbf{A},\, C_{t:t+T}^{\text{text}})
    \label{eq:interface}
\end{equation}
\end{tcolorbox}

\begin{figure}[t]
    \centering
    \includegraphics[width=1\linewidth]{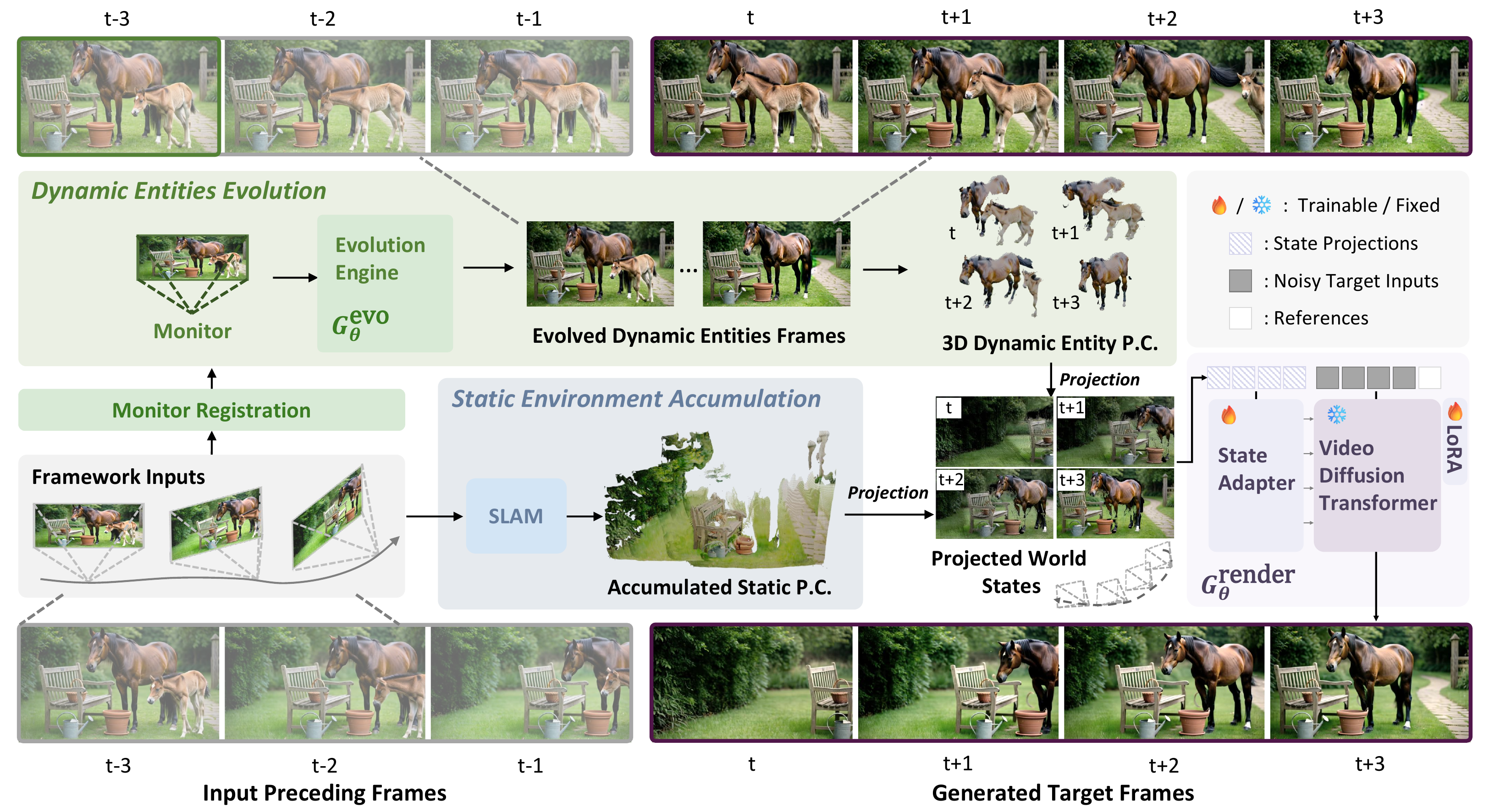}
    \vspace{-4.5mm}
    \caption{\textbf{LiveWorld overview.} Our system explicitly decouples world modeling into two processes. \textbf{(1) Static Accumulation (Blue):} Temporally-invariant backgrounds are fused into a static 3D point cloud via SLAM. \textbf{(2) Dynamic Evolution (Green):} Stationary monitors use the Evolution Engine $G_{\theta}^{\text{evo}}$ to fast-forward the out-of-sight progression of active entities, lifting them into 4D point clouds. \textbf{(3) State-aware Rendering (Purple):} Both representations are projected onto the target camera trajectory. This geometric projection, alongside appearance references, guides the renderer $G_{\theta}^{\text{render}}$ to synthesize coherent observations reflecting the elapsed dynamics.}
    \label{fig:main}
    \vspace{-2.5mm}
\end{figure}

\begin{figure}[t]
    \centering
    \includegraphics[width=1\linewidth]{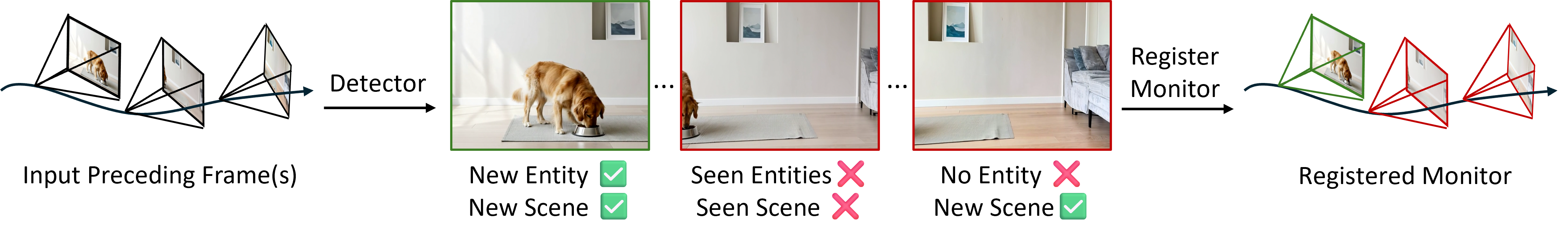}
    \vspace{-7mm}
    \caption{Given one or multiple preceding frames from the previous round, we first detect if the scene visited by the observer contains active dynamic entities, using off-the-shelf VLMs and segmentors. Following a positive detection, we further validate if the entity and scene are already registered by existing monitors.}
    \label{fig:monitor_registration}
    \vspace{-4mm}
\end{figure}

\vspace{-1mm}
\subsection{Evolving World States}
In this section, we introduce the maintenance of the world state formulated in Eq.~\ref{eq:world_state} through a continuous, multi-round update process. In each generation round spanning the temporal window $t$ to $t+T$, the system iteratively prepares the up-to-date world state $\mathcal{W}_{t:t+T}$ by executing two fundamental updates: \textbf{(1)} accumulating newly observed static regions into the temporally-invariant 3D representation $\mathcal{M}_{static}$, and \textbf{(2)} autonomously updating the registered active entities via the evolution function $G_{\theta}^\text{evo}$ to simulate their out-of-sight dynamics over this window into $\mathcal{M}_{dyn,t:t+T}$.
\vspace{-5mm}
\subsubsection{Accumulating the static environment $\mathcal{M}_{static}$}
We represent the static environment $\mathcal{M}_{static}$ as an accumulated background point cloud. For each historically observed frame in $\text{F}_{<t}$, the background is segmented~\cite{sam3} and incrementally fused into a global static point cloud via a feed-forward SLAM framework Stream3R~\cite{stream3r2025}, enabling continuous online memory accumulation.

\vspace{-3mm}
\subsubsection{Evolving dynamic entities $\mathcal{M}_{dyn}$}
\label{sec:event_evolution}
We implement the evolution function $\mathcal{E}$ through a \textit{monitor-driven dynamic evolution} system. Instead of modeling the entire unobserved world, we dynamically allocate stationary virtual agents, termed \textit{Monitors}, to track localized active regions.

% \noindent
\vspace{-2mm}
\paragraph{Defining and registering monitors.} Monitors are a set of generative agents placed at different world positions along the past camera trajectory, aiming to continuously evolve the previously observed scenes containing dynamic entities. They are powered by shared modules: a VLM-based detector~\cite{qwen3vl, sam3} for entity detection, and the evolution engine $G_{\theta}^\text{evo}$.
Specifically, as illustrated in Fig.~\ref{fig:monitor_registration}, before generating the new chunk $F_{t:t+T}$, we inspect the previously generated frames $F_{t-T:t}$ using the detector. If an unseen dynamic entity is detected along the observer's trajectory, and its spatial overlap with existing monitored regions falls below a certain threshold, a new monitor is registered at that specific pose (denoted as its \textit{anchor pose}). The monitor then takes the observed frame as its \textit{anchor frame}. To maintain computational efficiency, we limit the number of active monitors to $M$, discarding the one farthest from the observer when this limit is exceeded.

\vspace{-2mm}
% \noindent
\paragraph{Monitor-driven dynamic evolution.} After registration, we take the instantiated evolution engine $G_{\theta}^\text{evo}$ to simulate ongoing out-of-sight dynamics in the upcoming round for each monitor. First, to fix the monitor at the desired position without camera movements, recalling the state injection formulation (Eq.~\ref{eq:interface}), 
% we instantiate $G_{\theta}^\text{evo}$'s explicit state condition $\mathbf{P}_{t:t+T}$ with the repeated background of the anchor frame to be $\mathbf{P}^{\text{bg}_{anc}}_{t:t+T}$, simulating a static scene state with no background pixel change, forcing $G_{\theta}^\text{evo}$ to generated frames with static background. Second, to evolve foreground entities, we condition  $G_{\theta}^\text{evo}$ on text prompts $C_{t:t+T}^{text}$ detailing anticipated entity actions. For the supplementary appearance references, we cropped entities $\mathbf{A}^{\text{entity}}$ from the anchor frame and feed them into the DiT backbone. Passing these inputs through the shared backbone yields a synthesized local video of the entity's continued dynamics over the $t:t+T$ window.
We instantiate the explicit state condition $\mathbf{P}_{t:t+T}$ of $G_{\theta}^{\text{evo}}$ using the static background from the anchor frame, and condition $G_{\theta}^\text{evo}$ on text prompts $C_{t:t+T}^{text}$ detailing the anticipated actions of the entities to evolve the foreground entities, then cropped entities $\mathbf{A}^{\text{entity}}$ from the anchor frame for the additional appearance references. These inputs are processed through $G_{\theta}^{\text{evo}}$ to generate a local video depicting the continued dynamics of the entity over the interval $t:t+T$.

% \noindent
\vspace{-2mm}
\paragraph{Asynchronous temporal synchronization.} Since a new entity may emerge mid-round (e.g., first observed by the main camera at $t_a \in (t-T, t)$), its initial state is asynchronous with the current global timestamp $t$. To synchronize it before the next evolution round, the evolution engine $G_{\theta}^{\text{evo}}$ first synthesizes the missing local frames from $t_a$ to $t$. This aligns the entity's state with the global timeline, preparing it for the upcoming $t:t+T$ evolution.

\begin{tcolorbox}[width=1.0\linewidth, colframe=blackish, colback=lightgraybg, boxsep=0mm, arc=1mm, left=1mm, right=1mm, top=1mm, bottom=1mm]
\textbf{Instantiation I: Evolution Engine ($G_{\theta}^{\text{evo}}$).} We instantiate $G_{\theta}$ to synthesize the monitor's local video $v_{t:t+T}^{\text{monitor}}$, which is subsequently lifted to form $\mathcal{M}_{dyn, t:t+T}$. The inputs are specialized for localized event evolution:
\begin{equation}
    v_{t:t+T}^{\text{monitor}} = G_{\theta}^{\text{evo}}(\mathbf{P}^{\text{bg}_{anc}}_{t:t+T},\, \mathbf{A}^{\text{entity}},\, C_{t:t+T}^{\text{text}})
\end{equation}
where $\mathbf{P}^{\text{bg}_{anc}}_{t:t+T}$ is the repeated static anchor frame background, $\mathbf{A}^{\text{entity}}$ is the cropped entity reference, and $C_{t:t+T}^{\text{text}}$ is the action prompt.
\end{tcolorbox}

% \noindent
\vspace{-2.5mm}
\paragraph{Integrating dynamic memory.} Finally, equipped with the known monitor anchor pose and per-frame depth, the monitor unprojects the 2D dynamic foreground from $v_{t:t+T}^{\text{monitor}}$ back into the 3D world space. This lifting process yields a localized, temporally evolving 4D Monitor Point Cloud. This explicit 4D representation constitutes the concrete output $\mathcal{M}_{dyn, t:t+T}$ defined in Eq.~\ref{eq4:dynamic_memory}, providing the up-to-date dynamic memory to compose $\mathcal{W}_{t:t+T}$.

\vspace{-3mm}
\subsection{Rendering World States}
\vspace{-0.5mm}
\label{sec:observer_rendering}
With the updated world state $\mathcal{W}_{t:t+T}$ prepared (including the static point cloud $\mathcal{M}_{static}$ and the evolved $\mathcal{M}_{dyn, t:t+T}$), the final step is to synthesize the observer's visual experience. Here, we instantiate the unified backbone $G_{\theta}$ (Sec.~\ref{sec:shared_video_backbone}) into its second role: the state-aware renderer $G_{\theta}^\text{render}$. 

Unlike the evolution engine that operates on stationary monitor poses, $G_{\theta}^\text{render}$ synthesizes the scene from a continuously moving observer trajectory. Specifically, we project both the static 3D environment and the evolved dynamic monitor frames into the observer's novel camera views to construct the global state projection $\mathbf{P}_{t:t+T}$. Guided by this projection and historical reference frames, $G_{\theta}^\text{render}$ autoregressively renders the final observation video $F_{t:t+T}$, completing the generation loop.
\vspace{-3mm}
\subsubsection{Projecting the evolved world states}
% At each generation round spanning $t$ to $t+T$, the updated static point cloud $\mathcal{M}_{static}$ and the evolved dynamic 4D point clouds $\mathcal{M}_{dyn, t:t+T}$ are used to derive the explicit state projection $\textbf{P}_{t:t+T}$ for the state adapter. 
% Specifically, recalling the unified backbone configuration (Sec.~\ref{sec:shared_video_backbone}), to instantiate the renderer $G_{\theta}^{\text{render}}$ for generating the observation chunk $F_{t:t+T}$ under the camera trajectory $C_{t:t+T}^{\text{cam}}$, we project the composed world state $\mathcal{W}_{t:t+T} \approx \{\mathcal{M}_{static}, \mathcal{M}_{dyn, t:t+T}\}$ onto the target sequence of views:
At each generation round spanning $t$ to $t+T$, the updated static point cloud $\mathcal{M}_{static}$ and the evolved dynamic 4D point clouds $\mathcal{M}_{dyn, t:t+T}$ are used to derive the explicit state projection $\textbf{P}_{t:t+T}$ for the state adapter of $G_{\theta}^{\text{render}}$:
\vspace{-1mm}
\begin{equation}
    \mathbf{P}_{t:t+T} = \text{Proj}(\{\mathcal{M}_{static}, \mathcal{M}_{dyn, t:t+T}\}, C_{t:t+T}^{cam}) \in \mathbb{R}^{T \times H \times W \times C}
\end{equation}
As a result, generation strictly conditioned on the state projection $\mathbf{P}_{t:t+T}$ enables explicit camera control while consistently reflecting the spatial evolution of both the static environment and dynamic events throughout the sequence.

\vspace{-3mm}
\subsubsection{Reference frames retrieval for appearance guidance}
% To complete the rendering process, $G_{\theta}^{\text{render}}$ supplements the sparse geometric projection $\mathbf{P}_{t:t+T}$ with dense visual details via the Appearance LoRA. We retrieve the latest preceding frame $F_{t-1}$ as a temporal anchor for motion continuity and support image-to-video generation. Additionally, if $C_{t:t+T}^{cam}$ revisits previously explored regions, we retrieve the corresponding historical frames from $\text{F}_{<t}$ as appearance anchors. This strategy ensures high visual fidelity and texture consistency across continuous explorations.
To supplement dense visual details through the registered LoRA of $G_{\theta}^{\text{render}}$, we retrieve the latest preceding frame $F_{t-1}$ as a temporal anchor for motion continuity and if $C_{t:t+T}^{cam}$ revisits previously explored regions, we retrieve the corresponding eariliest historical frames, containing least visual drifting, from $\text{F}_{<t}$ as appearance anchors. This strategy ensures high visual fidelity and texture consistency across continuous explorations.

\begin{tcolorbox}[width=1.0\linewidth, colframe=blackish, colback=lightgraybg, boxsep=0mm, arc=1mm, left=1mm, right=1mm, top=1mm, bottom=1mm]
\textbf{Instantiation II: Observer Renderer ($G_{\theta}^{\text{render}}$).} We instantiate $G_{\theta}$ to render the global observation chunk $F_{t:t+T}$ based on the composed world state and the text control:
\vspace{-1mm}
\begin{equation}
    F_{t:t+T} = G_{\theta}^{\text{render}}(\mathbf{P}_{t:t+T}^{\text{global}},\, \mathbf{A}^{\text{history}},\, C_{t:t+T}^{\text{text}})
\end{equation}
where $\mathbf{P}_{t:t+T}^{\text{global}} = \text{Proj}(\{\mathcal{M}_{static}, \mathcal{M}_{dyn, t:t+T}\}, C_{t:t+T}^{\text{cam}})$ integrates both static and out-of-sight dynamics, and $\mathbf{A}^{\text{history}}$ provides retrieved spatial textures.
\end{tcolorbox}

\vspace{-3mm}
\subsection{Model Training}
\label{sec:training}

The unified backbone $G_{\theta}$ is trained with the flow matching objective~\cite{sd3}. Given a clean target latent $\mathbf{z}_0$ encoded by a frozen VAE, noise $\boldsymbol{\epsilon} \sim \mathcal{N}(0, \mathbf{I})$, and timestep $t \sim \mathcal{U}[0,1]$, the training loss is:
\vspace{-0.5mm}
\begin{equation}
    \mathcal{L} = \mathbb{E}_{\mathbf{z}_0, \boldsymbol{\epsilon}, t} \left\| \mathbf{v}_\theta(\mathbf{z}_t, t, \mathbf{P}, \mathbf{A}, C) - (\boldsymbol{\epsilon} - \mathbf{z}_0) \right\|^2
\end{equation}
where $\mathbf{z}_t = (1-t)\mathbf{z}_0 + t\boldsymbol{\epsilon}$. The loss is computed only on target frames; preceding and reference frames serve as clean conditioning tokens. We adopt a two-stage strategy: \textbf{Stage 1} trains the state adapter with the backbone frozen, and \textbf{Stage 2} freezes the adapter and fine-tunes LoRA modules on the backbone's attention layers for appearance reference integration. All encoders remain frozen throughout. Training data construction can be found in the Appendix.

%% file: secs/4_experiments.tex
\vspace{-1mm}
\section{Experiments}
\label{sec:exp}
\subsection{Introducing LiveBench}
\vspace{-0.5mm}
\subsubsection{Benchmark Construction.}
We aim to quantitatively evaluate the long-horizon maintenance of dynamic events by pairing diverse scene images with procedurally generated multi-round camera trajectories and text-driven scripts.
\vspace{-0.5mm}
\paragraph{Scene image curation.}
We curate 100 diverse scene images featuring various foreground entities and backgrounds. Using a VLM\cite{qwen3vl} for prompt composition and a text-to-image model \cite{qwenimage}, we generate photorealistic $480\times832$ images, strictly enforcing sharp focus, and clean, depth-unambiguous backgrounds.
\vspace{-0.5mm}
\paragraph{Trajectory design.}
After estimating scene geometry via Stream3R\cite{stream3r2025}, we generate camera trajectories that alternate between \emph{leaving} and \emph{revisiting} the initial viewpoint. We define two families: \textbf{(i) Same-Pose Revisit} ($A \to B \to A \to B \to A$ over four rounds), where the camera returns to its original pose, and \textbf{(ii) Different-Pose Revisit} ($A \to B \to C$), yielding revisits from novel viewpoints. We scale trajectories to fit individual scene sizes, generate 4 variants per scene (two families $\times$ left/right), each spanning 4 rounds (260 frames at 16 FPS).
% \vspace{-0.2cm}
\paragraph{Event scripts.}
A VLM generates per-step event scripts conditioned on the scene, ensuring physically plausible motions with explicit spatial displacements. In total, LiveBench comprises 100 scenes and 400 evaluation sequences.

\vspace{-1mm}
\subsubsection{Quantitative Evaluation Metrics.}
Due to the lack of ground-truth videos for out-of-sight dynamics, we design a reference-based and VLM-driven protocol.

\vspace{-1mm}
\paragraph{Spatial Memory and Identity.}
For same-pose revisits, we evaluate static background consistency against the initial frame (excluding dynamic regions) using PSNR, SSIM, and LPIPS~\cite{zhang2018lpips}. For dynamic entities, we compute the Chamfer Distance between the generated dynamic point clouds and the monitor's predictions in 3D world space. Additionally, across both revisit scenarios, we employ a Masked Bag-of-Features (BoF) strategy using foreground-masked DINOv2 tokens ($\text{DINO}_{fg}$) to robustly evaluate identity preservation under severe pose variations.
\vspace{-1mm}
\paragraph{Event Progression and Consistency.}
Since pixel metrics struggle with out-of-sight transitions, we utilize VLM-based binary VideoQA to verify if the generated actions and revisited states align with the text scripts (VQA-Acc). Finally, temporal smoothness of the evolved local dynamics is measured via adjacent-frame CLIP similarity ($\text{CLIP}_{F}$).

\subsection{Experimental Setup}

% \textbf{Training Data.}
% We curate ${\sim}$40k clips from SpatialVID~\cite{spatiavid} and RealEstate10K~\cite{realestate10k} at equal ratio, extracted at 16\,FPS and resized to $832{\times}480$. Each clip is processed through Qwen3-VL~\cite{qwen3vl} for entity detection, SAM3~\cite{sam3} for segmentation, and Stream3R~\cite{stream3r2025} for depth and pose estimation. Training samples comprise 65 target frames with 1 or 5 preceding frames (supporting both first-frame and multi-frame conditioning), up to 10 scene references selected by 2D coverage, and 10 foreground instance references. All projections are rendered from the accumulated point cloud and encoded with a frozen VAE.

\noindent
\textbf{Implementation Details.}
We build upon Wan2.1-14B-T2V~\cite{wan} with the state adapter initialized from Wan2.1-VACE-14B and rank-64 LoRA on the backbone attention layers. Stage~1 trains the state adapter for 10k steps; Stage~2 fine-tunes LoRA for 5k steps, both at lr $1\times10^{-4}$ with cosine decay. Global batch size is 16 across 16 NVIDIA H200 GPUs in bf16 FSDP. We set the maximum active monitors $M{=}3$ and drop text prompts with probability 0.2.

\noindent
\textbf{Other Evaluations.}
Beyond initial-frame revisits in LiveBench, we conduct human evaluation for complex scenarios involving late-appearing entities and concurrent out-of-sight dynamics. We summon a new foreground entity via text prompts mid-generation. Evaluators then assign binary scores across three criteria: Presence, Identity ($\text{Id.}_{fg}$), and Event Consistency. From these, we report two metrics: \textbf{Event Succ.} (an individual event satisfies all three criteria) and \textbf{Full Succ.} (both concurrent events succeed simultaneously), strictly measuring the capacity for persistent multi-event simulation.

\begin{table}[t!]
\centering
\footnotesize % 缩小基础字体
\setlength{\tabcolsep}{3pt} % 压缩列间距，给数字留出适当呼吸空间
\caption{Quantitative comparison on LiveBench. Background (bg) metrics measure spatial memory against the first scene frame, foreground (fg) metrics measure entity identity preservation, and VQA-Acc evaluates the success of out-of-sight event progression guided by the text prompt. Columns in blue highlight the \textbf{\textcolor{cyan}{second revisit}} performance. $\dagger$ denotes our implementation.}
\vspace{-2mm}
\label{tab:main_revisit}
\resizebox{\textwidth}{!}{
% 把 tabular 的右侧块从 *{3} 改为 *{2}，总列数从 19 减到 17
\begin{tabular}{l *{6}{c >{\columncolor{cyan!10}}c} | *{2}{c >{\columncolor{cyan!10}}c}}
\toprule
 & \multicolumn{12}{c|}{\textbf{Same Pose Revisit}} & \multicolumn{4}{c}{\textbf{Different Pose Revisit}} \\
\cmidrule(lr){2-13} \cmidrule(lr){14-17}
& \multicolumn{2}{c}{$\text{PSNR}_{bg}\uparrow$} & \multicolumn{2}{c}{$\text{SSIM}_{bg}\uparrow$} & \multicolumn{2}{c}{$\text{LPIPS}_{bg}\downarrow$} & \multicolumn{2}{c}{$\text{CD}_{fg}\downarrow$} & \multicolumn{2}{c}{$\text{DINO}_{fg}\uparrow$} & \multicolumn{2}{c|}{\text{VQA-Acc}$\uparrow$} & \multicolumn{2}{c}{$\text{DINO}_{fg}\uparrow$} & \multicolumn{2}{c}{\text{VQA-Acc}$\uparrow$} \\
\cmidrule(lr){2-3} \cmidrule(lr){4-5} \cmidrule(lr){6-7} \cmidrule(lr){8-9} \cmidrule(lr){10-11} \cmidrule(lr){12-13} \cmidrule(lr){14-15} \cmidrule(lr){16-17}
& 1st & 2nd & 1st & 2nd & 1st & 2nd & 1st & 2nd & 1st & 2nd & 1st & 2nd & 1st & 2nd & 1st & 2nd \\
\midrule
MG-2 & 16.321 & 16.131 & 0.512 & 0.502 & 0.565 & 0.629 & 6.631 & 7.429 & 0.335 & 0.198 & 7.737 & 5.012 & 0.230 & 0.122 & 5.111 & 4.132 \\
GC-1 & 17.637 & 16.012 & 0.571 & 0.523 & 0.421 & 0.572 & \underline{2.107} & 6.236 & \underline{0.527} & 0.262 & \underline{20.125} & 10.273 & \underline{0.475} & 0.191  & \underline{18.799} & 8.397 \\
Spatia$^{\dagger}$ & \textbf{20.132} & \underline{19.020} & \underline{0.672} & \underline{0.649} & \textbf{0.297} & \textbf{0.310} & 4.031 & \underline{5.122} & 0.440 & \underline{0.416} & 19.205 & \underline{14.655} & 0.392 & \underline{0.363} & 18.723 & \underline{13.212} \\
\midrule
\textbf{Ours} & \underline{20.071} & \textbf{19.983} & \textbf{0.679} & \textbf{0.650} & \underline{0.301} & \underline{0.330} & \textbf{0.068} & \textbf{0.135} & \textbf{0.760} & \textbf{0.721} & \textbf{59.063} & \textbf{54.620} & \textbf{0.691} & \textbf{0.632} & \textbf{52.829} & \textbf{49.478} \\
\bottomrule
\end{tabular}
}
% \vspace{-2mm}
\end{table}

\subsection{Main Results}
% In this chapter, we first compare the general ability of revisiting out-of-sight events within a long horizon on LiveBench (Tab.~\ref{tab:main_revisit}), and human evaluation for multiple event revisits with our own ablated baseline (Tab. ~\ref{tab:multi_event_user_eval}), and finally evaluate the ability of the Evolution Engine in long-horizon generation (Tab.~\ref{tab:evolution_engine}).

\subsubsection{Initial frame revisiting on LiveBench.}
\begin{figure}[t]
    \centering
    \includegraphics[width=1\linewidth]{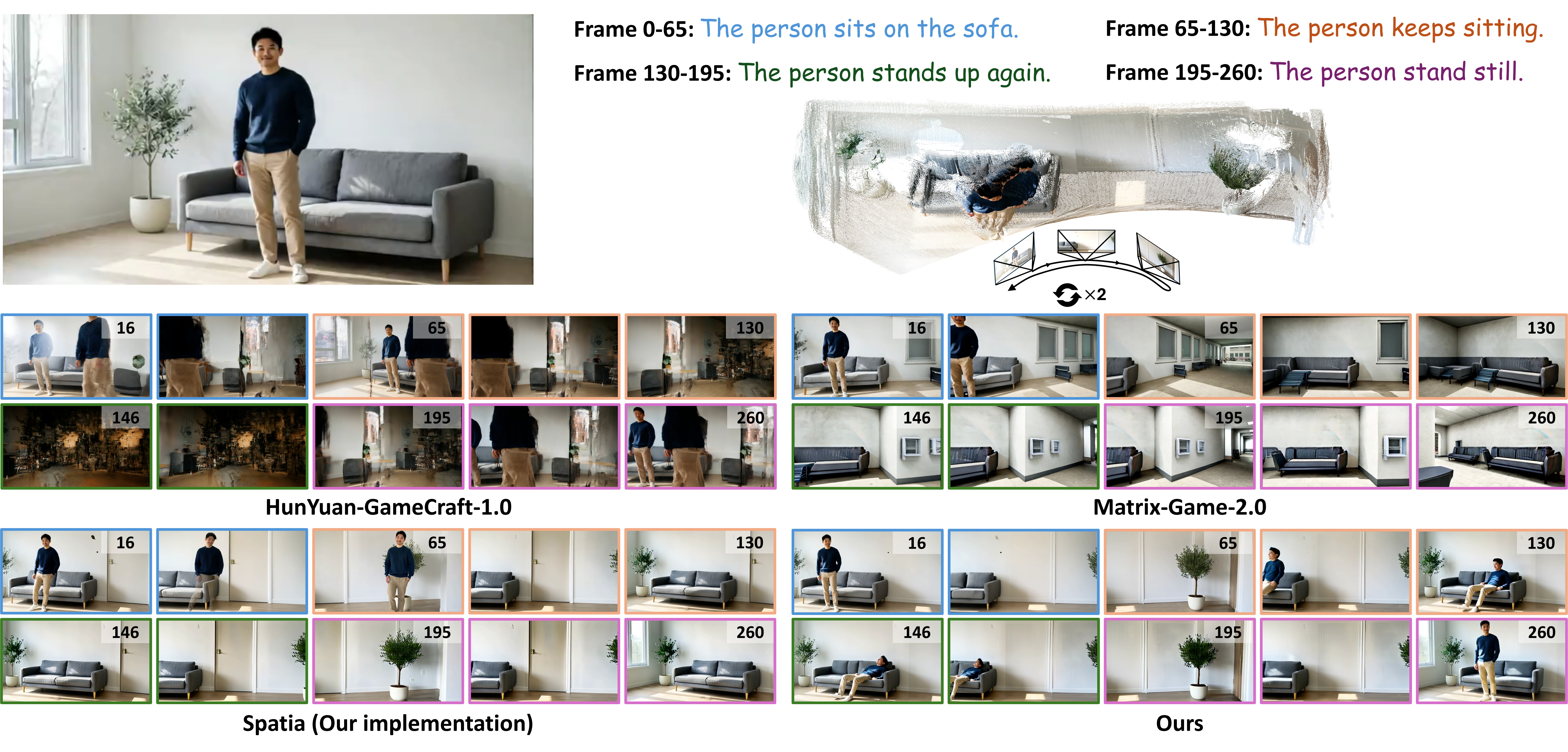}
    \vspace{-6mm}
    \caption{A comparison result with the latest state-of-the-art methods on LiveBench. With the camera view repeatedly moving rightwards and backwards, our methods stand out alone to successfully maintain long-horizon (260 frames) out-of-sight dynamics, while others fail. Different colors correspond to different evolving prompts of the event.}
    \label{fig:main_compare}
    \vspace{-2mm}
    
\end{figure}

Tab.~\ref{tab:main_revisit} presents the quantitative results for single-event revisiting. We choose state-of-the-art open-sourced camera-conditioned world models that support long horizon generation as comparison baselines, namely Matrix-Game-2.0 \cite{matrixgame}, Hunyuan-GameCraft-1.0 \cite{hunyuangamecraft}, and Spatia \cite{spatia}. We analyze the performance across four crucial dimensions:
\vspace{-2mm}

\paragraph{Spatial Background Maintenance.}
Benefiting from explicitly accumulated 3D point clouds, both our method and Spatia effectively maintain the static environment, achieving superior background metrics ($\text{PSNR}_{bg}$, $\text{SSIM}_{bg}$, $\text{LPIPS}_{bg}$). Without explicit spatial memory, Matrix-Game-2.0 (MG-2) and GameCraft-1 (GC-1) struggle during the first revisit. By the second long-horizon revisit (cyan columns), their backgrounds collapse completely with severe artifacts (Fig.~\ref{fig:main_compare}), causing a precipitous metric drop.
\vspace{-1mm}

\paragraph{Dynamic Entity Preservation.}
Crucially, LiveWorld uniquely preserves out-of-sight dynamic objects. Our decoupled Evolution Engine persistently updates world dynamics, while explicit state projection ensures perfect foreground alignment upon re-observation, achieving drastically better geometric ($\text{CD}_{fg}$) and semantic ($\text{DINO}_{fg}$) consistency. Baselines fundamentally fail here; despite history caching and prompt conditioning, they cannot guarantee foreground consistency across temporal gaps.
\vspace{-1mm}

\paragraph{Event Progression.}
Our decoupled architecture ensures highly successful text-script completion ($\text{VQA-Acc}$). By continuously evolving dynamic objects in the background, our renderer accurately captures their logically progressed states upon revisiting. Conversely, baselines entangle foreground motion and camera control within a single generator, causing camera movements to easily disrupt event generation and fail designated scripts.
\vspace{-1mm}

\paragraph{Different Pose Revisit.}
Our decoupled design's advantages amplify under novel revisiting viewpoints. While we maintain consistent entity identities ($\text{DINO}_{fg}$) and event alignment, baselines suffer further degradation due to exacerbated artifacts and failed camera control from novel angles.

% \begin{figure}[t]
%     \centering
%     \includegraphics[width=1\linewidth]{imgs/demo1.png}
%     \caption{main results place holder}
%     \label{fig:placeholder}
% \end{figure}

% --- 表 1：主实验（单事件重访系统级对比） ---

% --- 表 2：多事件重访人工评估 ---
\begin{table}[t!]
\centering
\footnotesize

\setlength{\tabcolsep}{8pt}
% \vspace{-3mm}% 稍微放大列间距，给数字呼吸空间
\caption{User study on late-appearing event revisit success rate, higher is better (\%).}
\label{tab:multi_event_user_eval}
% 用普通的 tabular 搭配外部缩放，彻底告别挤压
\vspace{-1mm}
\resizebox{0.78\textwidth}{!}{
\begin{tabular}{l *{9}{c}}
\toprule
\multirow{2}{*}{\textbf{Method}} & \multicolumn{2}{c}{\textbf{Presence}} & \multicolumn{2}{c}{\textbf{Id.}$_{fg}$} & \multicolumn{2}{c}{\textbf{Consistency}} & \multicolumn{2}{c}{\textbf{Event Succ.}} & \multirow{2}{*}{\makecell{\textbf{Full} \\ \textbf{Succ.}}} \\
\cmidrule(lr){2-3} \cmidrule(lr){4-5} \cmidrule(lr){6-7} \cmidrule(lr){8-9}
& E1 & E2 & E1 & E2 & E1 & E2 & E1 & E2 & \\
\midrule
w/o Event Evo & 40 & 39 & 32 & 31 & 20 & 15 & 2 & 3 & 0 \\
\textbf{Ours} & \textbf{92} & \textbf{70} & \textbf{80} & \textbf{61} & \textbf{67} & \textbf{53} & \textbf{42} & \textbf{35} & \textbf{26} \\
\bottomrule
\end{tabular}
}
% \vspace{-5mm}
\end{table}

\vspace{-2mm}
\subsubsection{Late-appear Event Revisiting.}
Since baselines struggle to initialize multiple entities via text, we benchmark against our camera-control variant (\textit{w/o Event Evo}). As Table~\ref{tab:multi_event_user_eval} shows, our method robustly maintains parallel events, achieving 92\% Presence for the primary event (E1). E2 Presence drops to 70\% as text-to-video randomness occasionally fails to trigger the entity, causing cascading metric declines. Nevertheless, our approach vastly outperforms the baseline with 42\% and 35\% \textbf{Event Succ.} for E1 and E2. Crucially, on the strict scene-level \textbf{Full Succ.} metric (requiring both events to succeed simultaneously), our model secures 26\% while the baseline completely collapses (0\%). This confirms that explicit evolution is indispensable for multi-event modeling.

% \subsection{Ablation Studies.}
% To validate the necessity of each core component, we conduct comprehensive ablation studies (Table~\ref{tab:ablation}).

% \noindent\textbf{Effect of Event Evolution (w/o Event Evo.).} 
% Removing the evolution engine effectively degrades our system into a pure camera control model. As expected, while it maintains competitive background scores, it fundamentally loses the ability to preserve out-of-sight dynamic entities. Its foreground preservation ($CD_{fg}$, $DINO_{fg}$) and event completion rate (VQA-Acc) drop drastically.

% \noindent\textbf{Effect of Spatial Memory (w/o Spa. Mem.).} 
% Disabling spatial memory leads to a catastrophic failure in camera control, essentially freezing the generation in place. Because the system still injects historical reference frames without proper spatial grounding, the attention LoRA incorrectly projects past reference content onto novel views. This misalignment causes severe ghosting and spatial jittering, degrading background metrics. 

% \noindent\textbf{Effect of Reference Frames (w/o Ref. Frame).} 
% Omitting historical reference frames deprives the model of dense visual textures, immediately destabilizing the generated background. This spatial instability accelerates the overall temporal collapse of the scene. This cascading failure is starkly reflected by the severe degradation across all metrics—especially in the foreground and event alignment scores—during the 2nd long-horizon revisit.

\vspace{-3mm}
\subsection{Ablation Studies}
To validate the necessity of each core component, we conduct comprehensive ablation studies in Tab.~\ref{tab:ablation}.

\vspace{-1mm}
\paragraph{Effect of Event Evolution.} Removing the evolution engine degrades our system into a pure camera control model. While background scores remain competitive, it completely fails to preserve out-of-sight entities, drastically dropping foreground ($CD_{fg}$, $DINO_{fg}$) and event completion (VQA-Acc) metrics.

\vspace{-1mm}
\paragraph{Effect of Spatial Memory.} Disabling spatial memory causes catastrophic camera control failure. Injecting historical references without proper spatial grounding misaligns the attention LoRA on novel views, resulting in severe ghosting, spatial jittering, and degraded background metrics.

\vspace{-1mm}
\paragraph{Effect of Reference Frames.}  Omitting historical references deprives the model of dense visual textures, destabilizing the background. This spatial instability triggers a cascading temporal collapse of the scene, severely degrading all metrics, especially foreground and event alignment, during the 2nd long-horizon revisit.

\begin{figure}[t!]
\vspace{-0.5mm}
    \centering
    \includegraphics[width=1\linewidth]{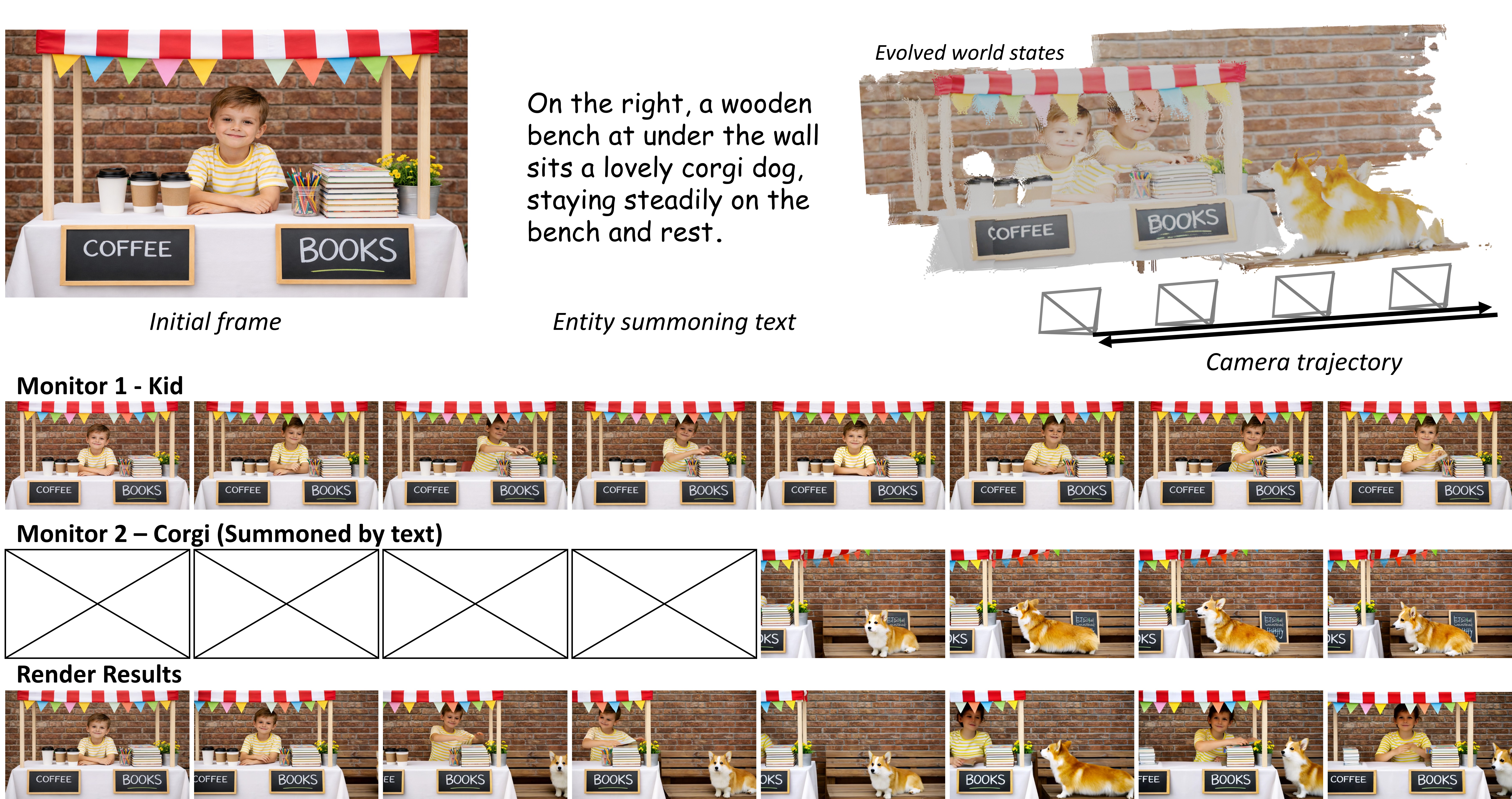}
    \caption{A demonstration of the late-appearing event revisiting. While we have the initial image containing the kid, we allow the renderer to generate the corgi following the text prompt. The monitor for the corgi is registered after the camera has overlap under the threshold with the monitor of the initial frame. The results showcase a perfect synchronization between the renderer and each monitor.}
    \label{fig:late_coming}
    % \vspace{-1mm}
\end{figure}

% --- 表 4：消融实验 ---
\begin{table}[t!]
\centering
\footnotesize
\setlength{\tabcolsep}{3pt}
\caption{Ablation studies of our proposed components. We validate the necessity of the evolution engine, spatial memory, and reference frames. The column in blue demonstrates the performance of a \textbf{\textcolor{cyan}{second revisit}}.}
\vspace{-1mm}
\label{tab:ablation}
\resizebox{\textwidth}{!}{
\begin{tabular}{l *{6}{c >{\columncolor{cyan!10}}c} | *{2}{c >{\columncolor{cyan!10}}c}}
\toprule
\multirow{3}{*}{\textbf{Method}} & \multicolumn{12}{c|}{\textbf{Same Pose Revisit}} & \multicolumn{4}{c}{\textbf{Different Pose Revisit}} \\
\cmidrule(lr){2-13} \cmidrule(lr){14-17}
& \multicolumn{2}{c}{$\text{PSNR}_{bg}\uparrow$} & \multicolumn{2}{c}{$\text{SSIM}_{bg}\uparrow$} & \multicolumn{2}{c}{$\text{LPIPS}_{bg}\downarrow$} & \multicolumn{2}{c}{$\text{CD}_{fg}\downarrow$} & \multicolumn{2}{c}{$\text{DINO}_{fg}\uparrow$} & \multicolumn{2}{c|}{\text{VQA-Acc}$\uparrow$} & \multicolumn{2}{c}{$\text{DINO}_{fg}\uparrow$} & \multicolumn{2}{c}{\text{VQA-Acc}$\uparrow$} \\
\cmidrule(lr){2-3} \cmidrule(lr){4-5} \cmidrule(lr){6-7} \cmidrule(lr){8-9} \cmidrule(lr){10-11} \cmidrule(lr){12-13} \cmidrule(lr){14-15} \cmidrule(lr){16-17}
& 1st & 2nd & 1st & 2nd & 1st & 2nd & 1st & 2nd & 1st & 2nd & 1st & 2nd & 1st & 2nd & 1st & 2nd \\
\midrule
w/o Event Evo.  & 20.005 & \textbf{18.995} & 0.671 & 0.647 & \textbf{0.299} & \textbf{0.315} & 4.215 & 5.431 & 0.425 & 0.401 & 18.512 & 13.985 & 0.380 & 0.350 & 18.105 & 12.855 \\
w/o Spa. Mem.   & 17.512 & 16.895 & 0.550 & 0.531 & 0.495 & 0.550 & 5.820 & 6.915 & 0.395 & 0.285 & 12.350 & 8.510  & 0.315 & 0.185 & 10.125 & 6.850 \\
w/o Ref. Frame  & 18.520 & 17.105 & 0.610 & 0.545 & 0.385 & 0.490 & 1.520 & 4.850 & 0.615 & 0.410 & 38.510 & 22.150 & 0.550 & 0.320 & 32.105 & 18.550 \\
\midrule
\textbf{Full} & \textbf{20.071} & 18.983 & \textbf{0.679} & \textbf{0.650} & 0.301 & 0.330 & \textbf{0.068} & \textbf{0.135} & \textbf{0.760} & \textbf{0.721} & \textbf{59.063} & \textbf{54.620} & \textbf{0.691} & \textbf{0.632} & \textbf{52.829} & \textbf{49.478} \\
\bottomrule
\end{tabular}
}
\vspace{-1mm}
\end{table}

% \vspace{-1mm}

% \section{Conclusion}
% We identify and formalize a fundamental limitation of existing generative video world models, the out-of-sight dynamics problem: unobserved regions are frozen at their last seen state, failing to capture ongoing events. To address this, we propose LiveWorld, a novel framework that explicitly decouples continuous world evolution from view-dependent rendering. By factorizing the environment into a temporally-invariant 3D background and explicitly evolving dynamic entities, LiveWorld makes global 4D evolution computationally tractable, overcoming the static-world assumption. Our monitor-centric pipeline uses virtual monitors to fast-forward unobserved entity dynamics and synchronize them with the global timeline, allowing the observer to maintain spatial coherence under explicit camera control in a persistently evolving world. We also introduce LiveBench, a benchmark for out-of-sight dynamics. Overall, LiveWorld establishes a principled foundation for long-term 4D world modeling, bridging the gap between static memorization and true dynamic simulation.

\vspace{-1mm}
\section{Conclusion}
We formalize the \textit{out-of-sight dynamics} problem in video world models, where unobserved regions incorrectly freeze at their last seen state. To overcome this, we propose LiveWorld, a framework that explicitly decouples continuous world evolution from view-dependent rendering. By factorizing the environment into a static 3D background and utilizing a monitor-centric pipeline to autonomously fast-forward unobserved active entities, LiveWorld achieves tractable 4D modeling. Along with LiveBench, our dedicated benchmark, LiveWorld bridges the gap between static 2D memorization and persistent 4D dynamic simulation.

%% file: secs/5_appendix.tex
\setcounter{page}{1}
\begin{center}
    {\Large \textbf{LiveWorld: Simulating Out-of-Sight Dynamics in Generative Video World Models}}\\
    \vspace{5mm}
    {\large Supplementary Materials}
\end{center}
\begin{figure}
    \centering
    \includegraphics[width=1\linewidth]{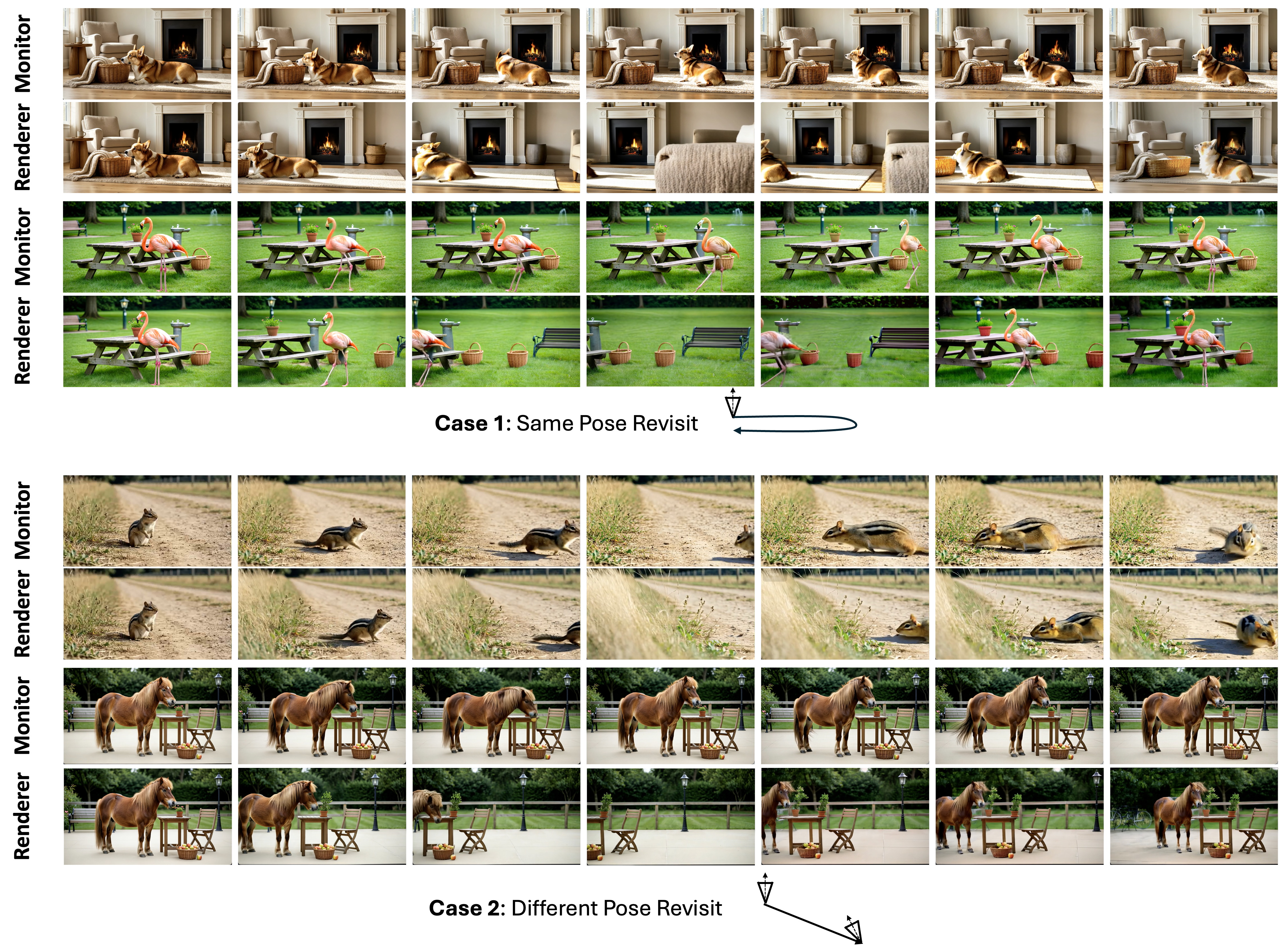}
    \caption{Typical revisit cases in our proposed LiveBench. More can be seen \href{https://zichengduan.github.io/LiveWorld/index.html}{here}.}
    \label{fig:more_cases}
\end{figure}
\section{Additional Results}
\subsection{Evaluating the evolution engine.}
Besides system-level effectiveness in Tab.~\ref{tab:main_revisit}, we additionally validate the performance of our unified model $G_{\theta}$ initialized as a $G_{\theta}^{\text{evo}}$, comparing with vanilla Wan2.1-I2V-14B (Wan) \cite{wan} in Tab.\ref{tab:evolution_engine}, evaluated on LiveBench scene images. Here, we compare the foreground entity script following (VQA-Acc), temporal frame consistency ($\text{CLIP}_{F}$), foreground ID preservation ($\text{DINO}_{fg}$), and background consistency ($\text{PSNR}_{bg}$). Despite Wan outperforming our finetuned $G_{\theta}^{\text{evo}}$ in the first round, our method is more capable of maintaining its performance for two reasons: (1) Our method leverages reference frames to effectively provide high-quality reference appearances to alleviate the error accumulation during long-horizon generation. (2) Wan sometimes struggles to maintain a static camera, leading to instability in pixel position shift, while ours maintains a completely static background movement thanks to the design introduced in Sec.\ref{sec:event_evolution}.

\subsection{More visualization results for different revisit cases.}
Fig.~\ref{fig:more_cases} shows more illustrations of the inference results for images in LiveBench, including both same pose revisits and different pose revisits.

% --- 表 3：演化器长时序评估 ---
\begin{table}[htbp]
\centering
\footnotesize
\setlength{\tabcolsep}{8pt} % 调大间距，让内容自然伸展
\caption{Evaluation of the Evolution Engine ($G_{\theta}^{\text{evo}}$) over long horizons. We compare round 1 (R1) and round 4 (R4) generations to demonstrate the robustness of our monitor-driven local evolution in maintaining script alignment, temporal coherence, identity, and background stability. The blue columns highlight the \textbf{\textcolor{cyan}{R4}} performance.}
\vspace{-3mm}
\label{tab:evolution_engine}
% 外部框死 0.8\textwidth，内部用原生 tabular 自适应
\resizebox{0.8\textwidth}{!}{
\begin{tabular}{l *{4}{c >{\columncolor{cyan!10}}c}}
\toprule
\multirow{2}{*}{\textbf{Method}} & \multicolumn{2}{c}{\text{VQA-Acc (\%)} $\uparrow$} & \multicolumn{2}{c}{$\text{CLIP}_{F}\uparrow$} & \multicolumn{2}{c}{$\text{DINO}_{fg}\uparrow$} & \multicolumn{2}{c}{$\text{PSNR}_{bg}\uparrow$} \\
\cmidrule(lr){2-3} \cmidrule(lr){4-5} \cmidrule(lr){6-7} \cmidrule(lr){8-9}
& R1 & R4 & R1 & R4 & R1 & R4 & R1 & R4 \\
\midrule
Wan-I2V \cite{wan} & \textbf{70.983} & 42.505 & \textbf{0.988} & 0.932 & \textbf{0.882} & 0.695 & \textbf{27.122} & 20.231 \\
\midrule
\textbf{Ours ($G_{\theta}^{\text{evo}}$)} & 61.774 & \textbf{50.541} & 0.977 & \textbf{0.972} & 0.811 & \textbf{0.794} & 26.481 & \textbf{22.639} \\
\bottomrule
\end{tabular}
}
\end{table}

\section{Training Dataset Construction}

Following \cite{epic}, we curate ${\sim}$40k 130-frame clips from MiraData~\cite{mira}, SpatialVID~\cite{spatiavid}, and RealEstate10K~\cite{realestate10k}, extracted at 16\,FPS and resized to $832{\times}480$. For MiraData, we additionally employ Qwen3-VL~\cite{qwen3vl} to filter clips with drastic scene changes, ensuring temporal continuity.

\paragraph{Entity Detection and Segmentation.}
We apply Qwen3-VL-8B-Instruct to identify up to 4 dynamic foreground entity categories (\textit{e.g.}, persons, vehicles) from a reference frame, filtering static background elements via a predefined blacklist. The detected categories serve as text prompts for SAM3~\cite{sam3}, which segments and propagates instance masks bidirectionally across all frames, producing per-frame binary foreground masks.

\paragraph{Scene Geometry Estimation.}
We employ Stream3R~\cite{stream3r2025} to estimate per-frame metric depth maps, camera intrinsics, and camera-to-world (c2w) poses with full attention across frames. Depth maps are upsampled to $832{\times}480$ via bilinear interpolation for denser point cloud construction.

\begin{figure}[t!]
    \centering
    \includegraphics[width=1\linewidth]{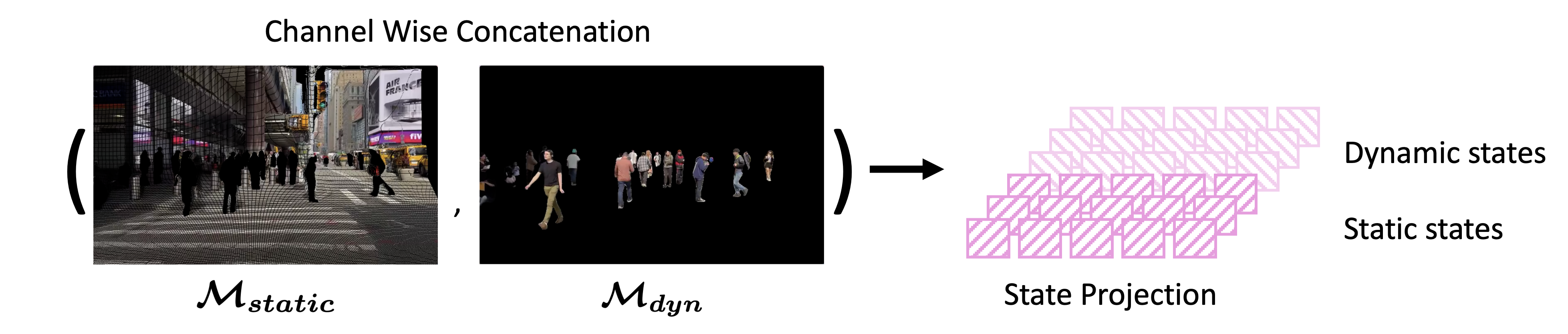}
    \caption{Demonsration of the ground truth frame of $\mathcal{M}_{static}$ and $\mathcal{M}_{dyn}$. The state adapter (in Sec.~\ref{sec:shared_video_backbone} and Fig.~\ref{fig:main}) takes the channel-wise concatenation of the $\mathcal{M}_{static}$ and $\mathcal{M}_{dyn}$ latents as input during training.}
    \label{fig:placeholder}
\end{figure}

\paragraph{Ground-Truth State Projection.}
To train the state adapter (Sec.~\ref{sec:shared_video_backbone}), we construct ground-truth state projections $\mathbf{P}_{t:t+T}$ following the structured world-state decomposition $\mathcal{W}_t \approx \{\mathcal{M}_{static},\, \mathcal{M}_{dyn,t}\}$.
For the \textbf{static component}, we build $\mathcal{M}_{static}$ by unprojecting valid depth pixels (${>}0$ and ${<}100$\,m) from selected reference frames into world coordinates, excluding SAM3 foreground regions to retain only temporally-invariant geometry. For each target and preceding frame, we render the static projection via Z-buffered perspective projection, producing projected RGB and validity masks.
For the \textbf{dynamic component}, we extract per-frame foreground appearance $\mathcal{M}_{dyn,t}$ by masking out the static background using the SAM3 masks and retaining only the foreground RGB.

\paragraph{State Projection Injection Strategy.}
The two projections are provided as separate channels to the state adapter, yielding a 32-channel latent input (16 per component). This design allows the model to freely combine static and dynamic signals. However, we observe that training with strictly separated channels leads to strong background control but weak foreground adherence, which we attribute to the relatively small spatial extent of foreground entities producing insufficient control signal. On the other hand, directly fusing $\mathcal{M}_{static}$ and $\mathcal{M}_{dyn,t}$ at the RGB level during training causes the model to over-rely on the projected pixels, losing the ability to hallucinate content in regions where the projection is incomplete (\textit{e.g.}, due to point-cloud holes caused by projection). As a practical compromise, we train with separated channels but overlay $\mathcal{M}_{dyn,t}$ onto $\mathcal{M}_{static}$ at inference time, providing a foreground prior in the static channel while retaining the flexibility of the independent dynamic channel. All experiments in this paper are conducted under this setting. We leave the exploration of more principled state projection injection mechanisms to future work.

\paragraph{Training Sample Assembly.}
Each sample comprises 65 target frames with 1 or 9 preceding frames, supporting both image- and clip-conditioned generation. Up to 10 scene reference frames $\mathbf{A}^{\text{history}}$ are selected via voxel-occupancy IOU ($\geq\!0.04$) with target frames to ensure geometric diversity, and up to 5 cropped foreground instance references $\mathbf{A}^{\text{entity}}$ are extracted with the same dot-dropout augmentation. All components---target frames, preceding frames, state projections, and references---are encoded with a frozen VAE~\cite{wan}, where reference frames are encoded per-frame independently to avoid unexpected temporal correlations.

\section{Benchmark Details}
\subsection{Benchmark Construction}
We construct LiveBench through a two-stage pipeline: (1) trajectory template generation per source image, and (2) assembly of inference-ready configurations combining entity detection, storyline generation, scene reconstruction, and trajectory plans.

\paragraph{Scene Reconstruction.}
Given a single source image, we estimate metric depth, camera intrinsics, and the camera-to-world pose using Stream3R~\cite{stream3r2025}. The depth map is unprojected into a world-coordinate point cloud with voxel downsampling at $0.01$\,m resolution.

\paragraph{Entity Detection and Storyline Generation.}
Dynamic foreground entities are detected using Qwen3-VL~\cite{qwen3vl} and filtered to retain only dynamic objects. A multi-step storyline of up to $S_{\max}{=}4$ evolution steps is then generated, each describing an explicit spatial displacement of the detected entities (\textit{e.g.}, ``the person walks from the bench toward the tree''). Each description is validated to contain explicit displacement verbs and reference visible scene elements.

\paragraph{Trajectory Generation.}
As described in Sec.~\ref{sec:exp}, we define two trajectory families scaled to individual scene sizes, each spanning 4 rounds (65 frames per round, 260 frames total) with left/right variants:
\begin{itemize}
  \item \textbf{Same-Pose Revisit} ($A{\to}B{\to}A{\to}B{\to}A$): lateral translation with simultaneous yaw rotation, where the camera returns to its original pose during the middle/last revisiting.
  \item \textbf{Different-Pose Revisit} ($A{\to}B{\to}C$): two independent translation stages with yaw alignment at each destination, yielding revisits from novel viewpoints.
\end{itemize}

To ensure perceptually consistent camera displacement across scenes of different scales, we adopt a \emph{screen-uniform} distance calibration. Instead of a fixed metric displacement, we target a consistent on-screen horizontal shift:
\begin{equation}
  d = \frac{w_{\text{img}} \cdot r_{\text{shift}} \cdot z_{\text{ref}}}{f_x},
\end{equation}
where $w_{\text{img}}{=}832$, $r_{\text{shift}}{=}0.18$ (18\% of frame width), $z_{\text{ref}}$ is the 45-th percentile depth of the scene point cloud, and $f_x$ is the focal length.
For trajectories involving lateral translation with foreground objects exiting (Case1 round1/3, Case2 round1), we additionally solve the translation amplitude such that the foreground object exits the camera view at approximately 75\% at such rounds, determined by projecting foreground calibration points into the camera frame.

\subsection{Benchmark Example}
Fig.~\ref{fig:bench_example} illustrates a concrete LiveBench instance. The source image depicts a person standing inside a warehouse. The benchmark produces two types of scripts that drive the renderer $G_{\theta}^{\text{render}}$ and the monitors $G_{\theta}^{\text{evo}}$ respectively:

\begin{tcolorbox}[width=1.0\linewidth, colframe=blackish, colback=lightgraybg, boxsep=0mm, arc=1mm, left=1mm, right=1mm, top=1mm, bottom=1mm]
\textbf{LiveBench Instance (\texttt{case1\_left}, Same-Pose Revisit).}

\vspace{1mm}
\begin{minipage}{\linewidth}
\centering
\includegraphics[width=0.9\linewidth]{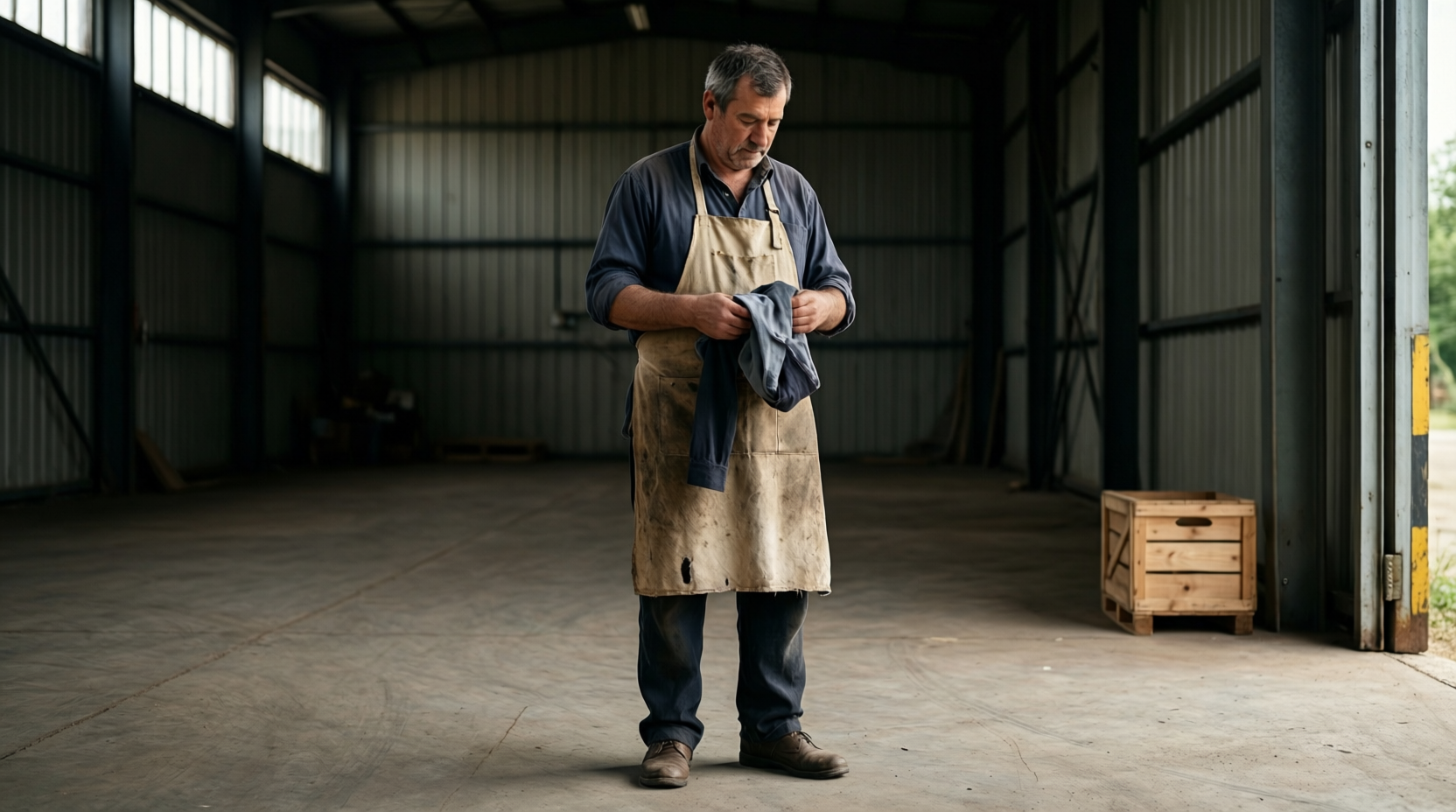}
\captionof{figure}{A person in a warehouse interior.}
\label{fig:bench_example}
\end{minipage}

\vspace{1mm}
\textbf{Renderer script} ($G_{\theta}^{\text{render}}$, per-round scene descriptions $C_{t:t+T}^{\text{text}}$): Controls the observer camera along the \emph{Same-Pose Revisit} trajectory ($A{\to}B{\to}A{\to}B{\to}A$), with per-iteration scene text:
\vspace{-1mm}
\begin{itemize}
  \item \textbf{R0}: \textit{``The wide, empty warehouse interior features corrugated metal walls and a concrete floor, with natural light filtering through high windows on the left. A wooden crate rests near the open doorway on the right, revealing greenery outside.''}
  \item \textbf{R1}: \textit{``The static environment remains coherent, with background layers and occlusion boundaries shifting smoothly along the trajectory.''}
  \item \textbf{R2--R3}: (Revisit rounds, analogous to R0--R1)
\end{itemize}

\vspace{1mm}
\textbf{Monitor script} ($G_{\theta}^{\text{evo}}$, per-step evolution prompts $C_{t:t+T}^{\text{text}}$): Drives the out-of-sight evolution of the detected entity (\textit{person}):
\vspace{-1mm}
\begin{itemize}
  \item \textbf{S0}: \textit{``The man walks toward the wooden crate near the doorway and pauses beside it.''}
  \item \textbf{S1}: \textit{``The man strolls toward the far wall, pausing near the metal beams.''}
  \item \textbf{S2}: \textit{``The man walks toward the wooden crate by the doorway and leans slightly against it.''}
  \item \textbf{S3}: \textit{``The man walks toward the far wall, strolls past the pallets, and stops near the corrugated metal beam.''}
\end{itemize}

\vspace{1mm}
\end{tcolorbox}
Additionally, the instance includes a file containing per-round camera poses, depth, and intrinsics for constructing $\mathbf{P}_{t:t+T}$; and an json file recording the detected dynamic entity categories.

\section{Discussions and Insights}
\subsection{Comparing representative world modeling benchmarks}
Alongside algorithmic advancements, several benchmarks have been recently proposed to evaluate video generation and world models. For instance, WorldScore \cite{worldscore} and WorldModelBench \cite{worldmodelbench} primarily focus on visual quality under specific camera trajectories and adherence to physical laws, while WorldSimBench \cite{worldsimbench} assesses general visual realism. A concurrent benchmark, MIND \cite{mind2026}, introduces a closed-loop evaluation specifically targeting memory consistency and action control. However, MIND explicitly defines memory consistency as ensuring that previously observed objects and layouts remain ``unchanged'' when revisited. This evaluation protocol inherently reinforces the static-world assumption, penalizing any natural temporal progression. In stark contrast, our LiveBench is distinctly designed to evaluate \textit{out-of-sight dynamics}. Instead of strictly matching the frozen historical snapshots, LiveBench's evaluation protocol rigorously assesses whether unobserved dynamic entities have naturally evolved over time upon revisiting, filling a critical void in benchmarking true 4D world simulation.

\subsection{More discussion with related concurrent works}
\label{sec:concurrent_work}
Apart from mainstream autoregressive video models and world models introduced in the related work, we also note a concurrent work, FloWM~\cite{lillemark2026flow}, which shares similar insights of modeling \textit{partially observed dynamic environments}---a concept that aligns with our formulation of \textit{out-of-sight dynamics}. While both works aim to prevent unobserved regions from being incorrectly frozen in memory, our methodologies and target scenarios \textbf{differ fundamentally}:

\paragraph{Latent Flow Extrapolation vs. Explicit Generative Evolution.}
FloWM tackles the problem from a representation learning perspective, enforcing Lie group flow equivariance within the latent space to extrapolate unobserved object trajectories. This mathematical constraint is effective for continuous, rigid physical motions (e.g., moving objects with constant velocities). In contrast, LiveWorld introduces an explicit, system-level decoupling mechanism. By grounding dynamic evolution through virtual monitors and state-conditioned video diffusion priors, we do not merely extrapolate spatial trajectories; \textbf{we simulate semantically rich, non-rigid, and discrete event progressions} (e.g., a person standing up, or an animal eating).

\paragraph{Open-World High-Fidelity Synthesis.}
While FloWM aims at maintaining latent state equivariance for dynamic tracking, LiveWorld is distinctly designed for high-fidelity generative video synthesis in the open world. Our structured world-state approximation seamlessly integrates temporally-invariant 3D SLAM backgrounds with autonomously evolving 4D foreground entities. This allows LiveWorld to handle complex appearance changes and maintain persistent visual textures across long-horizon explorations, bridging the gap between abstract state tracking and realistic visual rendering.

\subsection{Limitations and insights on future works}
As our work mainly focuses on providing a general formulation and a simple yet reasonable baseline, without hindering the contribution of our work and problem formulation, we would like to share several insights and observed drawbacks that can be concluded as starting points for future research in the community: 
\begin{itemize}
    \item[(1)] Our current system relies heavily on off-the-shelf models, including vision-language models serving as evolve engine scripts provider, segmentor serving as detector, and visual slam models for static point cloud aggregation; these additional module inevitably leads to system latency compared to end-to-end video generation models. Based on this, a clear direction could be to eliminate the need for explicit representations and turn to developing end-to-end models with implicit dynamic memories.
    \item[(2)] The quality of the dynamic foreground object rendering requires high-quality foreground point clouds, however, in the current implementation, we simply lift the monitor's video frames into 4D from its anchor pose, potentially leading to incomplete point clouds when the renderer revisits it from the other views. A possible solution would be to incorporate a monocular video-to-4D-mesh generator that provides dense, compact representations instead of the relatively sparsed point clouds, and thus provides consistently high-quality dense dynamic entity projections despite drastic view changes.
    \item[(3)] One would need to balance the inference time and the number of monitors in case of long generation latency. The precision of the approximation of the 4D world described in Eq.~\ref{eq:ideal_formulation} is also dependent on this balance accordingly.
    \item[(4)] Another interesting future direction would be to consider the interaction between the monitored scene, as currently our implementation regards as a minimal solution that satisfies the formulation, the interaction between the dynamic events in different monitors are yet to be designed.
    \item[(5)] Following the decoupled formulation in Eq.\ref{eq:ideal_formulation}, one may want to seek better representations of the world states (e.g., pre-defined point cloud priors \cite{pt3d}, latent world transition embeddings \cite{vjepa,vjepa2}, etc), and better models to render the observed world state \cite{ltx2}.
\end{itemize}

\section{Detailed System Designs}
% Supplementary Material — Inference Prompts and Subsystem Details
\subsection{Inference Prompts}
\label{sec:supp_inference_prompts}

We provide the complete set of prompts used during LiveWorld's inference pipeline. Our system relies on several language-model-driven modules that operate in concert: the \textbf{Observer Renderer} ($G_{\theta}^{\text{render}}$) for generating the observer's visual experience, the \textbf{Evolution Engine} ($G_{\theta}^{\text{evo}}$) for simulating out-of-sight dynamics at each monitor, and a suite of auxiliary subsystems including the \textbf{Entity Detector}, \textbf{Dynamic Filter}, and \textbf{Automatic Evolver} (Qwen3-VL-based prompt generator). Below, we detail each module's prompts and their roles.

% ----------------------------------------------------------------------------
\subsubsection{Renderer Prompts}
\label{sec:supp_renderer_prompts}
% ----------------------------------------------------------------------------

The Observer Renderer $G_{\theta}^{\text{render}}$ (Sec.~3.4 of the main paper) synthesizes the observer's visual experience by projecting the composed world state onto the target camera trajectory. Its text conditioning follows a structured two-part format that separately describes the static environment and the foreground entity dynamics.

\vspace{2mm}
\noindent\textit{Prompt composition format.}
The final text prompt fed to the renderer is composed from two independently generated components: a \texttt{scene\_text} describing the static environment, and an optional \texttt{fg\_text} describing foreground entity actions. These are composed as:

\begin{tcolorbox}[width=1.0\linewidth, colframe=blackish, colback=lightgraybg, boxsep=0mm, arc=1mm, left=2mm, right=2mm, top=2mm, bottom=2mm]
\textbf{Prompt Composition Rule.} Given \texttt{scene\_text} $S$ and \texttt{fg\_text} $F$:
\begin{itemize}
    \item If $F$ is non-empty: \texttt{"scene: \{S\}; foreground: \{F\}"}
    \item If $F$ is empty: \texttt{"scene: \{S\}"}
\end{itemize}
\end{tcolorbox}

\vspace{2mm}
\noindent\textit{Text prompts for exploring new static scenes.}
When the observer explores the environment, the renderer receives per-round \texttt{scene\_text} descriptions that characterize the static background visible from the current camera viewpoint. These descriptions are pre-generated by Qwen3-VL during \textbf{benchmark assembly}. They focus exclusively on the static environment---architecture, terrain, lighting, furniture, vegetation---and deliberately exclude any mention of dynamic foreground entities. Critically, the scene description prompt is \emph{trajectory-aware}: it receives a structured summary of the current camera state (position offset, yaw, motion phase) so that the VLM can imagine how the static environment appears from the interpolated viewpoint, even though it only observes the initial frame.

\begin{tcolorbox}[width=1.0\linewidth, colframe=blackish, colback=lightgraybg, boxsep=0mm, arc=1mm, left=2mm, right=2mm, top=2mm, bottom=2mm]
\textbf{Scene Description Prompt (for Qwen3-VL).} This prompt is used during benchmark assembly to generate the per-iteration \texttt{scene\_text} for the observer renderer. Given the source image, the list of foreground entities to exclude, and a trajectory state description, the VLM is asked to describe only the static environment:

\vspace{1mm}
\small
\texttt{You are labeling scene descriptions for an event-centric video benchmark. The source image shows the scene from the initial view. The current iteration corresponds to a different point along a pre-defined trajectory. Based on the image, IMAGINE and describe what the static environment would look like now.}

\vspace{1mm}
\texttt{Foreground entities to EXCLUDE (do NOT mention these at all): \{entities\}}

\vspace{1mm}
\texttt{Trajectory state for this iteration:}\\
\texttt{\{trajectory\_description\}}

\vspace{1mm}
\texttt{Rules:}\\
\texttt{- Describe only visible static-environment content in the current frame.}\\
\texttt{- Focus on layout/depth/occlusion changes as the viewpoint travels.}\\
\texttt{- Describe ONLY static environment: architecture, terrain, sky, vegetation, furniture, floor, walls, lighting, etc.}\\
\texttt{- Do NOT mention any animals, people, or dynamic foreground objects.}\\
\texttt{- Do NOT use control words like forward/backward/left/right/cw/ccw.}\\
\texttt{- Do NOT say ``camera faces/turns/points ...''.}\\
\texttt{- Keep it concise (1-2 sentences).}\\
\texttt{- Output the scene\_text string only, no JSON wrapping.}
\end{tcolorbox}

\noindent The \texttt{\{trajectory\_description\}} placeholder is filled with a structured camera state computed from the pre-planned trajectory, for example:

\begin{tcolorbox}[width=1.0\linewidth, colframe=blackish, colback=lightgraybg, boxsep=0mm, arc=1mm, left=2mm, right=2mm, top=2mm, bottom=2mm]
\textbf{Example Trajectory State} (injected into the scene description prompt):

\vspace{1mm}
\small
\texttt{- Motion action: leave; path segment: A->B}\\
\texttt{- Segment phase: 1/1}\\
\texttt{- Position offset (X,Z): (0.15, -0.03)}\\
\texttt{- Segment translation magnitude: 0.18}\\
\texttt{- Segment yaw change magnitude: 25.3 deg}\\
\texttt{- Current yaw value: -12.7 deg}\\
\texttt{- This is the beginning of the segment}
\end{tcolorbox}

\noindent An example of a generated \texttt{scene\_text} for a park scene:

\begin{tcolorbox}[width=1.0\linewidth, colframe=blackish, colback=lightgraybg, boxsep=0mm, arc=1mm, left=2mm, right=2mm, top=2mm, bottom=2mm]
\textbf{Example Scene Text (Round 0).}

\vspace{1mm}
\small
\texttt{A wooden bench with vertical slats occupies the left foreground, its armrest partially framing the view. Behind it, a stone wall and a flagpole with a red flag stand near a terracotta planter overflowing with pink flowers, all set against a backdrop of a brick building and green lawn.}
\end{tcolorbox}

\noindent At runtime, when benchmark assembly is not used (\ie, running in open-ended mode without pre-computed configs), an alternative simpler prompt is used to generate the \texttt{scene\_text} directly from the current anchor frame:

\begin{tcolorbox}[width=1.0\linewidth, colframe=blackish, colback=lightgraybg, boxsep=0mm, arc=1mm, left=2mm, right=2mm, top=2mm, bottom=2mm]
\textbf{Runtime Scene Description Prompt (for Qwen3-VL).} Used when generating \texttt{scene\_text} on-the-fly during inference (for both the renderer and the monitor):

\vspace{1mm}
\small
\texttt{Describe only the static environment visible in this image in 1-2 concise sentences. EXCLUDE any foreground entities (\{entities\}) --- do NOT mention them. Focus on architecture, terrain, lighting, furniture, walls, floor, vegetation, etc. Do NOT use camera motion words (forward/backward/left/right/pan/tilt). Output the description string only, no JSON wrapping.}
\end{tcolorbox}

\vspace{2mm}
\noindent\textit{Text prompts for summoning new foreground objects.}
LiveWorld supports \emph{late appearing entity summoning}: a user can introduce a new foreground entity mid-generation via a text prompt. In this case, the renderer receives both a \texttt{scene\_text} and a \texttt{fg\_text} that together form the composed prompt. The \texttt{fg\_text} is provided by the user or generated by the Automatic Evolver (see Sec.~\ref{sec:supp_auto_evolver}). For example, to summon a corgi into a coffee shop scene (as in Fig.~6 of the main paper):

\begin{tcolorbox}[width=1.0\linewidth, colframe=blackish, colback=lightgraybg, boxsep=0mm, arc=1mm, left=2mm, right=2mm, top=2mm, bottom=2mm]
\textbf{Example Entity Summoning Prompt.}

\vspace{1mm}
\small
\texttt{scene: On the right, a wooden bench at under the wall sits a lovely corgi dog, staying steadily on the bench and rest.}

\vspace{1mm}
The renderer generates the scene with the newly summoned entity, and the detector subsequently registers a new monitor for the corgi if it has sufficient spatial overlap with an existing monitor's coverage region.
\end{tcolorbox}

% ----------------------------------------------------------------------------
\subsubsection{Monitor Prompts (Evolution Engine)}
\label{sec:supp_monitor_prompts}
% ----------------------------------------------------------------------------

Each registered monitor drives the Evolution Engine $G_{\theta}^{\text{evo}}$ to simulate out-of-sight dynamics. The evolution engine is conditioned on a text prompt that describes how the detected entities should continue their natural motion. This prompt is generated by the \textbf{Automatic Evolver}, a Qwen3-VL-based module that observes the monitor's anchor frame and produces a physically plausible continuation description.

The monitor's text prompt follows the same composition format as the renderer: \texttt{"scene: \{scene\_text\}; foreground: \{fg\_text\}"}, where \texttt{scene\_text} is generated once per monitor (since the monitor camera is stationary) and \texttt{fg\_text} is re-generated each evolution round.

\begin{tcolorbox}[width=1.0\linewidth, colframe=blackish, colback=lightgraybg, boxsep=0mm, arc=1mm, left=2mm, right=2mm, top=2mm, bottom=2mm]
\textbf{Instantiation I: Evolution Engine ($G_{\theta}^{\text{evo}}$).} We instantiate $G_\theta$ to synthesize the monitor's local video $v_{t:t+T}^{\text{monitor}}$, which is subsequently lifted to form $\mathcal{M}_{dyn,t:t+T}$. The inputs are specialized for localized event evolution:
\vspace{-1mm}
\begin{equation}
    v_{t:t+T}^{\text{monitor}} = G_{\theta}^{\text{evo}}(\mathbf{P}_{t:t+T}^{\text{bg\_anc}},\, \mathbf{A}^{\text{entity}},\, C_{t:t+T}^{\text{text}})
\end{equation}
where $\mathbf{P}_{t:t+T}^{\text{bg\_anc}}$ is the repeated static anchor frame background, $\mathbf{A}^{\text{entity}}$ is the cropped entity reference, and $C_{t:t+T}^{\text{text}}$ is the action prompt. The action prompt $C_{t:t+T}^{\text{text}}$ is composed as: \texttt{"scene: \{scene\_text\}; foreground: \{fg\_text\}"}.
\end{tcolorbox}

\noindent An example of a complete monitor prompt for a scene containing a person:

\begin{tcolorbox}[width=1.0\linewidth, colframe=blackish, colback=lightgraybg, boxsep=0mm, arc=1mm, left=2mm, right=2mm, top=2mm, bottom=2mm]
\textbf{Example Monitor Prompt (Evolution Round 1).}

\vspace{1mm}
\small
\texttt{scene: A wooden bench with a plaque rests on a paved walkway beside a stone planter holding pink flowers and a wicker picnic basket. In the background, a stone building with arched windows and a flagpole stands under a clear sky, framed by green grass and trees.}

\vspace{1mm}
\texttt{foreground: The person walks toward the wooden bench, leans slightly against its backrest, and remains fully inside the frame.}
\end{tcolorbox}

% ============================================================================
\subsection{Prompts for Other Subsystems}
\label{sec:supp_subsystem_prompts}
% ============================================================================

% ----------------------------------------------------------------------------
\subsubsection{Entity Detector}
\label{sec:supp_entity_detector}
% ----------------------------------------------------------------------------

The Entity Detector is responsible for identifying active dynamic entities in the observer's field of view. It operates as a two-stage pipeline: (1)~\textbf{Qwen3-VL}~\cite{qwen3vl} performs open-vocabulary entity detection, and (2)~\textbf{SAM~3}~\cite{sam3} segments each detected entity at the instance level. Below we detail the prompts and operation of each stage.

\vspace{2mm}
\noindent\textbf{Stage 1: Scene-level entity detection (Qwen3-VL).}
Given the observer's preceding frames $F_{t-T:t}$, we select a representative frame and feed it to Qwen3-VL along with a carefully structured detection prompt. The prompt is designed to enumerate all non-static, potentially dynamic categories present in the scene, following a ``remove-what-is-not-background'' strategy:

\begin{tcolorbox}[width=1.0\linewidth, colframe=blackish, colback=lightgraybg, boxsep=0mm, arc=1mm, left=2mm, right=2mm, top=2mm, bottom=2mm]
\textbf{Scene Detection Prompt (Qwen3-VL).} This prompt is provided alongside the observed frame to identify foreground entities that should be tracked by monitors:

\vspace{1mm}
\small
\texttt{List CATEGORIES to remove, keeping ONLY static solid background.}

\vspace{1mm}
\texttt{BACKGROUND (keep): walls, floor, ceiling, furniture, doors, windows, buildings, roads, sidewalks, trees, grass, scenery, fixed structures.}

\vspace{1mm}
\texttt{REMOVE (even if stationary):}\\
\texttt{- People (including hands/arms/feet)}\\
\texttt{- Vehicles (car, truck, bus, motorcycle, bicycle, scooter, wheelchair)}\\
\texttt{- Animals}\\
\texttt{- Handheld/portable items (bag, phone, cup, food, umbrella)}\\
\texttt{- Movable objects (stroller, cart, luggage, carried boxes)}\\
\texttt{- Water surfaces (river, lake, ocean, pond, fountain, waterfall, stream, waves, puddle)}\\
\texttt{- Other dynamic elements (fire, flame, flags, signs, balloons, etc.)}\\
\texttt{- Sky (if visible in the scene, write exactly: sky)}

\vspace{1mm}
\texttt{RULES:}\\
\texttt{1. Output CATEGORIES, not individual instances}\\
\texttt{2. Use AT MOST 4 categories total (merge aggressively)}\\
\texttt{3. Pick broad categories that cover the most area}\\
\texttt{4. For any human, ALWAYS write exactly: person}\\
\texttt{5. For any sky region, ALWAYS write exactly: sky}\\
\texttt{6. For any water surface, ALWAYS write exactly: water}\\
\texttt{7. Keep items short (1-5 words)}\\
\texttt{8. When unsure, LIST IT (over-remove is better than under-remove)}

\vspace{1mm}
\texttt{OUTPUT FORMAT:} \texttt{Nothing} \texttt{OR numbered list (max 5 items):}\\ \texttt{1) person} \texttt{2) car} \texttt{3) sky}

\vspace{1mm}
\texttt{EXAMPLES:}\\
\texttt{Scene: office with desks, a person typing, coffee cup on desk} \texttt{1) person} \texttt{2) cup on the table}\\
\texttt{---}\\
\texttt{Scene: street with buildings, parked cars, two pedestrians, blue sky above}\\
\texttt{1) person} \texttt{2) car} \texttt{3) sky}\\
\texttt{---}\\
\texttt{Scene: riverside park, jogger running, river flowing}\\
\texttt{1) person} \texttt{2) water} \texttt{3) sky}\\
\texttt{---}\\
\texttt{Scene: empty room with table and chairs}\\
\texttt{Nothing}\\
\texttt{---}\\
\texttt{Scene: kitchen, person's hands visible preparing food}\\
\texttt{1) person} \texttt{2) cup on the table}
\end{tcolorbox}

\vspace{2mm}
\noindent\textbf{Stage 1.5: Dynamic entity filtering (Qwen3-VL).}
After the initial detection, we apply a second Qwen3-VL pass to filter out entities that cannot move on their own (\eg, a cup on a table detected in Stage~1). This ensures that only truly dynamic entities---people, animals, vehicles---are registered as monitors:

\begin{tcolorbox}[width=1.0\linewidth, colframe=blackish, colback=lightgraybg, boxsep=0mm, arc=1mm, left=2mm, right=2mm, top=2mm, bottom=2mm]
\textbf{Dynamic Filter Prompt (Qwen3-VL).} Given the same frame and the list of detected entities from Stage~1:

\vspace{1mm}
\small
\texttt{Given this image and these detected objects: \{entities\}}

\vspace{1mm}
\texttt{Which of these can move ON THEIR OWN? (people, animals, birds, vehicles, robots, etc.)}

\vspace{1mm}
\texttt{STATIC (exclude): furniture, appliances, fixtures, radiators, doors, windows, lamps, plants, decorations, buildings, structures, signs, sky, water.}

\vspace{1mm}
\texttt{List ONLY the dynamic/movable ones. If none can move, output exactly: Nothing}

\vspace{1mm}
\texttt{Output format:}\\
\texttt{Nothing}\\
\texttt{OR numbered list:}\\
\texttt{1) person}\\
\texttt{2) dog}
\end{tcolorbox}

\vspace{2mm}
\noindent\textbf{Stage 2: Instance-level segmentation (SAM~3).}
After filtering, each surviving entity category name (\eg, ``person'', ``dog'') is passed as a text prompt to SAM~3~\cite{sam3} for open-vocabulary instance segmentation. SAM~3 returns per-instance binary masks for all instances of that category in the frame. Each instance mask is then used to:
\begin{enumerate}
    \item Compute a tight bounding box with padding;
    \item Crop the entity from the original frame and place it on a black square background (preserving aspect ratio);
    \item Produce a cropped binary mask aligned with the cropped image.
\end{enumerate}
These cropped entity images serve as the appearance references $\mathbf{A}^{\text{entity}}$ for the Evolution Engine, and the masks are used for foreground--background separation during point cloud lifting.

\vspace{2mm}
\noindent\textbf{Intermediate detection (optional).}
For long-horizon generation, an optional intermediate detection module can be enabled to detect new entities that appear mid-sequence. This uses a simplified detection prompt:

\begin{tcolorbox}[width=1.0\linewidth, colframe=blackish, colback=lightgraybg, boxsep=0mm, arc=1mm, left=2mm, right=2mm, top=2mm, bottom=2mm]
\textbf{Intermediate Detection Prompt (Qwen3-VL).} Applied to generated frames during the observer's exploration to detect late-appearing entities:

\vspace{1mm}
\small
\texttt{Look at this frame and list any dynamic entities (people, animals, vehicles, or other objects that can move) that are clearly visible.}

\vspace{1mm}
\texttt{RULES:}\\
\texttt{1. Output CATEGORIES, not individual instances}\\
\texttt{2. Use AT MOST 3 categories total}\\
\texttt{3. For any human, ALWAYS write exactly: person}\\
\texttt{4. Keep items short (1-5 words)}\\
\texttt{5. Only list entities that are clearly visible and prominent}

\vspace{1mm}
\texttt{OUTPUT FORMAT:}\\
\texttt{Nothing}\\
\texttt{OR numbered list:}\\
\texttt{1) person}\\
\texttt{2) dog}
\end{tcolorbox}

\vspace{2mm}
\noindent\textbf{Entity deduplication.}
To prevent registering duplicate monitors for the same physical entity observed from different viewpoints or at different times, we employ DINOv2~\cite{dinov2} embeddings for visual similarity matching. When a new entity is detected, its cropped image embedding is compared against all existing monitor instances. If the cosine similarity exceeds a threshold $\tau_{\text{sim}} = 0.45$, the detection is considered a re-observation of an existing entity and is merged into the corresponding monitor rather than spawning a new one.

% ----------------------------------------------------------------------------
\subsubsection{Automatic Evolver}
\label{sec:supp_auto_evolver}
% ----------------------------------------------------------------------------

The Automatic Evolver generates physically plausible motion continuation prompts (\texttt{fg\_text}) for each monitor's evolution round. It operates in two modes: (1)~\textbf{pre-computed storyline} mode, where prompts are generated during benchmark assembly via \texttt{assemble\_event\_bench.py} and stored in \texttt{storyline.json}; and (2)~\textbf{on-the-fly} mode, where prompts are generated at runtime by conditioning Qwen3-VL on the monitor's current anchor frame. In both modes, the core system prompt is the same, but the benchmark assembly pipeline additionally appends \emph{hard motion rules} and a \emph{continuation chain} mechanism to ensure high-quality, diverse, and physically plausible motion descriptions.

\vspace{2mm}
\noindent\textbf{Base evolution system prompt.}
The following system prompt template is shared across both modes. It is stored in \texttt{system\_config.yaml} and parameterized by the detected entity names:

\begin{tcolorbox}[width=1.0\linewidth, colframe=blackish, colback=lightgraybg, boxsep=0mm, arc=1mm, left=2mm, right=2mm, top=2mm, bottom=2mm]
\textbf{Evolution System Prompt (Qwen3-VL).} This prompt is provided alongside the monitor's current anchor frame to generate the \texttt{fg\_text} for the next evolution round. The \texttt{\{entities\}} placeholder is filled with the detected entity names (\eg, ``person'', ``dog''):

\vspace{1mm}
\small
\texttt{You are a scene evolution agent for one fixed event camera.}\\
\texttt{The current scene contains these dynamic entities: \{entities\}.}\\
\texttt{Write one concise but vivid image-to-video continuation prompt describing how these entities continue moving naturally and coherently in THIS scene only.}

\vspace{2mm}
\texttt{MOTION GUIDELINES:}\\
\texttt{- Entities should INTERACT with objects visible in the scene. Move TOWARD a specific object or landmark (a bench, a table, a rock, a tree, a ball, a flower pot, a doorway, etc.) and interact with it (approach, sit on, lean against, sniff, circle around, settle beside, etc.).}\\
\texttt{- Good examples: a dog walks toward the bench and sits beside it; a person takes up the coffee cup and drink; a man stands up.; a cat pads toward the rock and settles on it; a child walks toward the swing set.}\\
\texttt{- ONLY gentle, steady motions: walking, strolling, trotting, padding, wandering, taking up.}\\
\texttt{- FORBIDDEN intense motions: running, dashing, jumping, leaping, hopping, sprinting, lunging, pouncing, chasing, bolting, galloping. These cause unstable video.}\\
\texttt{- Secondary motions (turning, gesturing, looking around, wagging tail) should accompany the larger movement, not replace it.}

\vspace{2mm}
\texttt{CRITICAL CONSTRAINT:}\\
\texttt{- ALL entities MUST stay fully inside the frame at ALL times. Never let any entity walk out of frame, move to the edge, or leave the visible area.}\\
\texttt{- Entities should move WITHIN the visible scene area, not toward or past its boundaries.}

\vspace{2mm}
\texttt{RULES:}\\
\texttt{- Keep motion physically plausible --- natural walking, strolling pace.}\\
\texttt{- Preserve identities.}\\
\texttt{- Do not add new objects or new identities.}\\
\texttt{- Return only the final prompt text.}
\end{tcolorbox}

\vspace{2mm}
\noindent\textbf{Hard motion rules (benchmark assembly).}
During benchmark assembly, the base system prompt is augmented with additional hard constraints to ensure that the generated motion descriptions contain sufficient displacement and object interaction. These rules are appended after the system prompt:

\begin{tcolorbox}[width=1.0\linewidth, colframe=blackish, colback=lightgraybg, boxsep=0mm, arc=1mm, left=2mm, right=2mm, top=2mm, bottom=2mm]
\textbf{Motion Hard Rules} (appended to the system prompt during storyline generation):

\vspace{1mm}
\small
\texttt{CRITICAL MOTION RULES (must follow):}\\
\texttt{- Prioritize INTERACTION with visible objects in the scene. Entities should move TOWARD a specific object or location (a bench, a table, a rock, a tree, a ball, a food bowl, a doorway, etc.) and interact with it (approach, sniff, sit on, lean against, circle around, settle beside, etc.).}\\
\texttt{- If no obvious object is nearby, the entity should travel to a clearly different area of the scene (from one side to the other, toward a visible landmark, etc.).}\\
\texttt{- Describe clear displacement AND a purpose/destination, not aimless wandering.}\\
\texttt{- Explicitly state WHERE the entity moves (e.g., ``walks toward the bench'', ``trots over to the flower pot'', ``moves from the left side to the table on the right'').}\\
\texttt{- ONLY gentle, steady motions: walking, strolling, trotting, padding, wandering. FORBIDDEN intense motions: running, dashing, jumping, leaping, hopping, sprinting, lunging, pouncing, chasing, bolting, galloping. These cause unstable video.}\\
\texttt{- FORBIDDEN local-only motions: turning head, raising/lowering head, nodding, glancing, looking around, tail wagging, tiny posture shifts, standing still.}\\
\texttt{- Each entity must end in a visibly different position than where it started.}\\
\texttt{- Keep all entities fully inside frame at all times. Never leave the visible area.}\\
\texttt{- Output 1-2 concise sentences only.}
\end{tcolorbox}

\vspace{2mm}
\noindent\textbf{Continuation chain mechanism.}
For multi-step storyline generation, each step after the first receives the previous step's output as context, along with explicit instructions to ensure diversity and progression:

\begin{tcolorbox}[width=1.0\linewidth, colframe=blackish, colback=lightgraybg, boxsep=0mm, arc=1mm, left=2mm, right=2mm, top=2mm, bottom=2mm]
\textbf{Continuation Chain Suffix} (appended for steps $>$0):

\vspace{1mm}
\small
\texttt{Previous evolution description: \{prev\_prompt\}}\\
\texttt{Continue naturally from where the previous evolution ended. The entities must move to a DIFFERENT object or location in the scene --- approach a bench, walk toward a tree, trot over to a ball, head to a doorway, etc. Prioritize INTERACTING with visible objects in the scene rather than aimless wandering. The viewer should see them end up in a visibly different spot than where they started. Do NOT let them stay in the same place, stand still, or only fidget. IMPORTANT: all entities must stay fully inside the frame at all times --- never walk out of frame or move to the edge. Do not repeat the previous description.}

\vspace{1mm}
\texttt{Write one compact motion sentence with explicit destination and frame safety, for example:}\\
\texttt{'<entity> walks from the left side toward the bench on the right while staying fully inside the frame.'}\\
\texttt{'<entity> strolls over to the flower pot near the wall and sniffs it, remaining inside the frame.'}
\end{tcolorbox}

\vspace{2mm}
\noindent\textbf{Automatic rewriting.}
The assembly pipeline validates each generated \texttt{fg\_text} against a set of regex-based rules that check for: (1)~out-of-frame motion references, (2)~stationary or in-place-only movement, (3)~local micro-motions without whole-body travel, (4)~missing travel verbs, (5)~missing displacement cues (start-to-destination), and (6)~missing in-frame safety phrases. If any check fails, Qwen3-VL is re-prompted with the violation reason appended (\eg, \texttt{``Your previous draft was invalid because it describes stationary or tiny in-place movement. Rewrite and strictly satisfy all motion rules.''}). This retry loop runs up to 3 attempts before falling back to a template-based motion description.

\vspace{2mm}
\noindent\textbf{Example storyline output.}
The complete generated storyline is stored as \texttt{storyline.json} and contains both the shared \texttt{scene\_text} and per-step \texttt{fg\_text}:

\begin{tcolorbox}[width=1.0\linewidth, colframe=blackish, colback=lightgraybg, boxsep=0mm, arc=1mm, left=2mm, right=2mm, top=2mm, bottom=2mm]
\textbf{Example Storyline} (\texttt{storyline.json}) for a scene containing a person near a park bench:

\vspace{1mm}
\small
\texttt{scene\_text}: \texttt{A wooden bench with a plaque rests on a paved walkway beside a stone planter holding pink flowers and a wicker picnic basket. In the background, a stone building with arched windows and a flagpole stands under a clear sky, framed by green grass and trees.}

\vspace{1mm}
\textbf{Step 0:} \texttt{The person strolls toward the wicker basket on the pavement, pauses beside it, and gently rests one hand on its edge while staying fully inside the frame.}

\vspace{1mm}
\textbf{Step 1:} \texttt{The person walks toward the wooden bench, leans slightly against its backrest, and remains fully inside the frame.}

\vspace{1mm}
\textbf{Step 2:} \texttt{The person strolls toward the wicker basket on the pavement, pauses beside it, and gently rests one hand on its rim, remaining fully inside the frame.}

\vspace{1mm}
\textbf{Step 3:} \texttt{The person strolls from the basket toward the wooden bench, leans lightly against its backrest, and remains fully inside the frame.}
\end{tcolorbox}

\noindent When the preset storyline mode is enabled (\texttt{use\_preset\_storyline: true}), the pre-computed \texttt{storyline.json} is loaded and the per-step \texttt{fg\_text} values are used directly as monitor prompts, bypassing runtime Qwen3-VL calls entirely. When disabled, the Automatic Evolver generates prompts on-the-fly using only the base system prompt (without the hard rules and continuation chain), conditioning Qwen3-VL on the monitor's latest anchor frame at each evolution round. This dual-mode design allows both fully automatic open-ended evolution and controlled, reproducible benchmark evaluation.

\vspace{2mm}
\noindent\textbf{Fallback prompts.}
When Qwen3-VL is unavailable (\eg, due to memory constraints or CPU-offload failures), the system falls back to template-based descriptions:

\begin{tcolorbox}[width=1.0\linewidth, colframe=blackish, colback=lightgraybg, boxsep=0mm, arc=1mm, left=2mm, right=2mm, top=2mm, bottom=2mm]
\textbf{Fallback Scene Text.}

\vspace{1mm}
\small
\texttt{The static environment remains coherent, with background layers and occlusion boundaries shifting smoothly along the trajectory.}

\vspace{2mm}
\textbf{Fallback Foreground Text} (parameterized by entity names and step index):

\vspace{1mm}
\small
\texttt{\{entities\} walk toward a nearby object in the scene, moving a clear distance from their current position to interact with it while staying fully inside the frame. (evolution step \{step+1\}/\{max\_steps\})}
\end{tcolorbox}